# Post-Inference Model Chaining: A Problem Statement and a Model-based Bootstrap for Training End-to-End Digital Twins

Lukas Madenach*^, Timo Bailer*, Tobias Eilert*°^

* CMC Statistics BioPharma, Boehringer Ingelheim Pharma GmbH & Co. KG, Ingelheim, Germany

° Corresponding author

^These authors contributed equally.

## Abstract

End-to-end digital twins can be constructed by fitting separate stage models and connecting them after inference. We investigate the Frequentist prediction properties of a commonly used Monte Carlo simulation scheme that forwards complete predictive draws through such a chain when measured stage outputs contain independent normal errors, but the latent stage means are deterministic. Analytical two-stage examples show persistent asymptotic interval and coverage inflation and, with quadratic propagation, mean bias dependent on the parameters to be estimated. Simulations of four- and eight-stage linear–quadratic univariate chains with 7, 15 or 30 training runs demonstrate pronounced input-dependent distortions. A diagnostic using error-free latent training predictors shows that the effect is mainly not attributable to errors in variables, but to the MCSim approach even showing underestimation of coverage. We propose a model-based bootstrap: an initial chain forwards fitted means, bootstrap chains are refitted on fixed-center bootstrap data while forwarding their mean prediction, and a target-stage future error enters a studentized prediction root. Across the investigated scenarios, relative mean bias is substantially reduced, and coverage is closer to the nominal 90%. At sample size 30, the eight-stage bootstrap has mean width ratios of 1.16–1.42 relative to the known-parameter normal reference, compared with 4.11–4.97 for observed-to-observed MCSim. These results support improved endpoint prediction in the tested setting. Further investigations of the bootstrap scheme in the supplement for a bivariate 6-stage process show similar satisfactory results.

## 1 Introduction

As industry moves towards greater automation and digitalization, digital twins — virtual representations of physical processes — are becoming essential tools for monitoring, control, steering and optimization [1, 2, 3]. While digital twins are increasingly implemented at individual process stages, their true potential lies in predicting how changes at one stage affect later stages and ultimately the final product. However, constructing a single, comprehensive model that captures the entire process, i.e., an end-to-end digital twin (E2E-DT), is often infeasible due to the system's complexity, data heterogeneity, data scarcity, computational constraints, and the need to model different stages with different model types. A pragmatic alternative is to model each stage separately using available input and output data and then reconnect these models sequentially also termed the modular approach [4, 5]. The statistical assessment distinguishes bias of the mean predictor, pointwise coverage for one new measured observation and prediction-interval width relative to a known-parameter reference [6]. Biopharmaceutical production motivates this work, but the results in this manuscript apply to all fields in which separately trained models of whichever type need to be chained post-inference to yield a complete model with approximate proper statistical properties for example in metrology [7, 8].

The main task in the modular approach is how to reconnect the proper single models so that the entire process is described by one model that satisfies the classical frequentist statistical properties. We do not discuss the Bayesian or the GUM approach via the law of propagation of uncertainty [9, 10]. We call this task holistic process modelling. In the literature there is one popular approach known for measurement error propagation, which uses Monte Carlo simulations abbreviated here as MCSim [11, 12, 13, 14, 15]. The idea is simple: given a future input for the first stage, the first stage model is utilized to sample from its predictive distribution, i.e., we obtain a collection of samples representing the uncertainty of the output. These samples are forwarded as inputs to the second model and for each input sample one output sample is drawn from the predictive distribution of the second model. This is now repeated until the last stage model. The lastly produced samples (or also for some stage in between) are then used to compute a mean prediction and quantiles that serve as prediction intervals. We examine the bias, coverage and width of this particular predictive-draw propagation scheme. The conclusions concern how fitted predictive distributions are connected after inference, not the numerical validity of Monte Carlo integration.

The following illustration distinguishes variability of a physical quantity from measurement error around a fixed underlying quantity. Both are represented by random variables, but their prediction targets differ. Composing the known physical variables gives the distribution of a transformed physical quantity; applying the same operation to a distribution around an estimate need not yield an unbiased estimate of the underlying transformed value. For example, let us assume we have a solution of spherical nanoparticles whose distribution of radii $R$ is ***known*** to be normal, i.e. $R \sim \mathcal{N}(\mu, \sigma)$ with known mean radius $\mu$ and standard deviation $\sigma$ (ignoring, for simplicity, the fact that radii cannot be negative). We emphasize again that there is no uncertainty here, only a stochastic process of random radii which is completely ***known*** since its type of probability distribution and its parameters are known. Let us assume that we are now interested in the distribution of the surface area of the nanoparticles giving rise to the random variable

$$S = 4\pi R^2 = 4\pi(\mu + \sigma z)^2 = 4\pi(\mu^2 + 2\mu\sigma z + \sigma^2 z^2),$$

where we used that $R = \mu + \sigma z$ with a standard normal random variable $z$. The expected value is

$$E(S) = 4\pi\mu^2 + 4\pi\sigma^2,$$

because of its linearity as well as $E(z) = 0$ and $E(z^2) = 1$. We see that we have an offset of $4\pi\sigma^2$ to the result of $4\pi\mu^2$. This is the impact of the randomness in our stochastic process of random radii on the average surface area, but not a "mistake"! Since the composition of random variables is effectively an integration, Monte Carlo simulations are the perfect tool here since they are a class of numerical methods to solve integrals.

We now change the target from a varying radius to a fixed but unknown radius measured with error. The issue is whether directly transforming a distribution around its estimate preserves the desired target, rather than whether Monte Carlo can evaluate the transformation numerically. As an introductory example, let us now assume a different situation: we still have nanoparticles in a solution, but they all have **exactly the same** radius $\mu$, which is ***unknown***. However, the analytical method to measure the radius has a normally distributed error with (for simplicity here) known standard deviation $\sigma$. If we now measure once, i.e. the measured radius is given by the random variable $R = \mu + \sigma z$, the uncertainty distribution estimator for the mean radius $\mu$ is $\hat{\mu} = R - \sigma z'$, where $z$ and $z'$ are independently standard normally distributed [16, 17]. Following the MCSim approach analytically, we plug the predictive uncertainty distribution estimator $\hat{\mu}$ into the surface area formula to get

$$\hat{S} = 4\pi\hat{\mu}^2 = 4\pi(R - \sigma z')^2 = 4\pi(\mu + \sigma z - \sigma z')^2 = 4\pi\left(\mu + \sqrt{2}\sigma\bar{z}\right)^2$$

$$= 4\pi\left(\mu^2 + 2\sqrt{2}\mu\sigma\bar{z} + 2\sigma^2\bar{z}^2\right),$$

with $\bar{z}$ standard normally distributed, and the expectation value becomes

$$E(\hat{S}) = 4\pi\mu^2 + 8\pi\sigma^2.$$

Here, the offset of $8\pi\sigma^2$ is not correct anymore, because it is obvious that the true surface area of any of the nanoparticles is $S = 4\pi\mu^2$. The ordinary squared plug-in estimator already has a nonlinearity-induced offset; adding and transforming an independent uncertainty draw doubles that offset. Thus, the additional distortion comes from the statistical construction, not from Monte Carlo integration itself. We will elaborate the effect of the MCSim approach on other statistical metrics in the context of holistic process modelling with more realistic examples in much more detail in Section 3.

For post-inference model chaining, the objective is therefore to quantify uncertainty for a clearly specified endpoint target rather than merely to compose fitted predictive distributions. Stage estimators may be dependent because they share measured training variables. Fitting the stages separately does not make these random estimators independent, but propagating independent future predictive errors through fixed fitted models does not reproduce their joint sampling distribution. A sequential bootstrap can instead repeat the fitted-chain estimator on surrogate datasets, making its sources of variation explicit. Whether that surrogate approximation is adequate remains a model-specific question.

However, even before the task of chaining, we are already faced with hurdles in the estimation of the single models. First, transformed predictors can become nearly collinear or yield numerically rank-deficient stage designs. Second, it is inherent to the problem that the input data for at least the second and the following stages are subject to random measurement errors. Near collinearity can make coefficient estimates unstable, but usually does not affect prediction performance drastically. The second point is known as the error-in-variables problem; it is more severe and needs to be dealt with [18]. Measurement error in the predictors can bias ordinary regression fits even when the latent stage model is correctly specified; subsequent nonlinear transformations can compound the resulting discrepancy. The latent-input diagnostic in this study isolates this issue from the separate consequences of the predictive propagation rule for the investigation of the MCSim approach.

The problem statement first specifies the deterministic latent chain and measured stage outputs. The investigation then develops analytical intercept-to-linear and intercept-to-quadratic examples, followed by simulations of four- and eight-stage chains alternating between linear and quadratic functions. The same simulation design includes observed-to-observed and latent-to-observed training for MCSim, so that predictor measurement error is not conflated with post-inference propagation.

The proposed alternative is a fixed-center studentized model-based bootstrap. Its initial estimator uses fitted preceding-stage means as predictors, and its pseudo-responses retain fixed initial stage centers while the bootstrap designs vary. A separate target-stage future error supplies the predictive component of the studentized root. We evaluate the resulting endpoint mean bias, prediction coverage and interval width on the same underlying four- and eight-stage processes as for the MCSim scheme. Additionally, we investigate its performance on a bivariate 3-stage chain in the supplement.

# 2 Problem statement and Notation

Industrial production processes are inherently iterative, i.e., consisting of consecutive steps, referred to hereafter as unit operations (UOs) or stages. Since the number of unit operations, say $n$, usually range up to 8-10, the complexity of each UO is high and the data is usually scarce, it is often not feasible or advisable to estimate one model describing the whole production line at once. A common workaround is that each unit operation is inferred separately and then post-hoc re-connected to yield a model describing the complete production process.

The first step to understand and then overcome the problems that arise in this approach is to define a statistical model for the production process further denoted by $F$. We start with the description of a mean production line originating from mean functional models for each unit operation $i$ given by the functions $\bar{f}_i: \mathbb{R}^{n_{i-1}} \longrightarrow \mathbb{R}^{n_i}$ each having $n_{i-1}$ real inputs and $n_i$ real outputs. The complete mean production line is then given by the composition of the single unit operations $\bar{F} := \bar{F}_{1,..,n} := \bigcirc_{i=1}^{n} \bar{f}_i: \mathbb{R}^{n_0} \longrightarrow \mathbb{R}^{n_n}$, where we used the convention

$$\left(\bigcirc_{i=1}^{n} \bar{f}_i\right)(x_0) = \left(\bar{f}_n \circ \bar{f}_{n-1} \circ \cdots \circ \bar{f}_2 \circ \bar{f}_1\right)(x_0) = \bar{f}_n\left(\bar{f}_{n-1}\left(\cdots \bar{f}_2\left(\bar{f}_1(x_0)\right)\right)\right),$$

for a starting configuration $x_0 \in \mathbb{R}^{n_0}$. After each UO, the output material is measured in many different attributes so that we add a homoscedastic centered error vector $\epsilon_i$ to the mean functions to account for this. Thus, the statistical model for the unit operation $i$ can be expressed by $f_i(x_{i-1}) = \bar{f}_i(x_{i-1}) + \epsilon_i$ for all $x_{i-1} \in \mathbb{R}^{n_{i-1}}$. The measurement errors are not propagated to the next unit operation; they are added to the measured output only. This results in a statistical model for the complete production line constructed as

$$F(x_0) = \bar{F}_{1,..,n}(x_0) + \varepsilon_n = \left(\bigcirc_{i=1}^{n} \bar{f}_i\right)(x_0) + \varepsilon_n.$$

Note that this statistical model represents the truth, which is ***not*** known to the engineers in reality. We assume that the true process is constructed from deterministic functions, i.e., we restrict ourselves to measurement errors and do not consider process variability at the unit operations. Measurement errors are always present, and separating them from process variability keeps the propagation of pure measurement uncertainty clear. However, correct propagation of statistical uncertainty including process variability is part of forthcoming work.

Given that the $k$-th execution of the process, also called a run or batch, starts at a starting configuration $x_0^k \in \mathbb{R}^{n_0}$ for $k \in \{1, \dots, m\}$, we in total collect $m$ data sets, where $m$ is the number of executions of the production line. These consist of independent random vectors defined by $X_i^k := F_{1,..,i}^k(x_0^k) := \left(\bigcirc_{j=1}^{i} \bar{f}_j\right)(x_0^k) + \varepsilon_i^k$, i.e., the random outputs of each unit operation given that we started at $x_0^k$, building up the random data vector $X^k := \left({x_0^k}^T, {X_1^k}^T, {X_2^k}^T, \cdots, {X_n^k}^T\right)^T$ for execution $k \in \{1, \dots, m\}$ and yielding the random data matrix $X = (X^1, \dots, X^m)^T$. The error vectors $\varepsilon_i^k$ are assumed to be iid between executions, i.e., independent copies of $\varepsilon_i$, as well as independent between unit operations. From each realization of the random data submatrix $X_i := \left(\left(X_{i-1}^1, X_{i-1}^2, \dots, X_{i-1}^m\right)^T, \left(X_i^1, X_i^2, \dots, X_i^m\right)^T\right)^T$, we have information to build a predictive estimator $\tilde{f}_i$ for the statistical model $f_i$ of the $i$-th unit operation [19]. An important observation is that the estimators of consecutive unit operations $i-1$ and $i$ share the random data $\left(X_i^1, X_i^2, \dots, X_i^m\right)^T$. This means that the estimator of a unit operation depends at least on the directly preceding unit operation, but usually also on earlier ones.

Now the ultimate goal in holistic process modelling is to estimate $F$, i.e., the entire production line, by one estimator, say $\tilde{F}$. Since this is not feasible, the post-inference approach needs to somehow use the collection of trained models $\{\tilde{f}_i\}_{i\in\{1,\dots,n\}}$ to obtain a suitable approximation of $\tilde{F}$. The MCSim approach takes the intuitive way of chaining these distribution estimators to obtain an approximation by

$$\tilde{F}_{MCSim}(x_0) := \left(\circ_{i=1}^n \tilde{f}_i\right)(x_0).$$

The next section investigates this approach in a frequentist setting.

# 3 Investigation of MCSim Approach

## 3.1 Analytical Univariate Case

As an example – that can be solved purely analytically – we look at a production line with only two simplified unit operations which have a univariate input as well as univariate output, i.e., $n_0 = n_1 = n_2 = 1$. The first mean function is set to be an intercept model, that is $\bar{f}_1(x_0) = a_1$, which is independent of the starting value $x_0$. The random output data points are then given by $X_1^k = f_1^k(x_0) = a_1 + \varepsilon_1^k$ with independent and normally distributed measurement errors $\varepsilon_1^k \sim N(0, \sigma_1)$ and positive and finite standard deviation $\sigma_1$. The random data is denoted by $X_1 = \{X_1^1, \dots, X_1^m\}$. It follows that a predictive distribution estimator for $f_1$ is given in the form

$$\tilde{f}_1(x_0) = \bar{X} + t_{m-1} S \sqrt{1 + \frac{1}{m}}$$

with $t_{m-1} \sim T_{m-1}$, the $t$-distribution with $m-1$ degrees of freedom,

$$\bar{X} = \frac{1}{m}\sum_{k=1}^{m} X_1^k = \frac{1}{m}\sum_{k=1}^{m} f_1^k(x_0) \sim N\left(a_1, \sigma_1/\sqrt{m}\right)$$

and

$$S^2 = \frac{1}{m-1}\sum_{k=1}^{m}\left(X_1^k - \bar{X}\right)^2 = \frac{1}{m-1}\sum_{k=1}^{m}\left(f_1^k(x_0) - \bar{X}\right)^2 = \sigma_1^2\theta/(m-1).$$

The latter two are the estimators for the sample mean and the unbiased sample variance of the random data. Here, the random variable $\theta$ follows a $\chi^2$-distribution with $m-1$ degrees of freedom. It is readily seen that the distribution estimator $\tilde{f}_1(x_0)$ is consistent, unbiased and shows correct coverage behavior for $f_1(x_0)$.

As a first holistic example, we choose an affine function for the second unit operation that is $f_2(x_1) = a_2 + b_2 x_1 + \varepsilon_2$ with $\varepsilon_2 \sim N(0, \sigma_2)$, $\sigma_2 > 0$ and $b_2 \neq 0$, where we assume that the coefficients are known, i.e., $\tilde{f}_2 = f_2$. This is not realistic, but it is useful to derive an analytic solution, and it is a simplification of the real situation. Given the true intercept model of the first unit operation and the true affine function of the second UO, the true holistic mean function is $\bar{F}(x_0) = a_2 + b_2 a_1$. Applying the MCSim approach analytically we obtain

$$\tilde{F}_{MCSim}(x_0) = \tilde{f}_2\left(\tilde{f}_1(x_0)\right) = f_2\left(\tilde{f}_1(x_0)\right) + \varepsilon_2$$

$$= a_2 + b_2\left(\bar{X} + t_{m-1} S \sqrt{1 + \frac{1}{m}}\right) + \varepsilon_2$$

$$= a_2 + b_2 \bar{X} + b_2 t_{m-1} S \sqrt{1 + \frac{1}{m}} + \varepsilon_2.$$

The target of the prediction is a new observation $F(x_0) = \bar{F}(x_0) + \varepsilon_2$ with target mean $\bar{F}(x_0) = E\big(F(x_0)\big) = a_2 + b_2 a_1$. Since $E(t_{m-1}) = 0$ (the draw of $t_{m-1}$ is independent of the data), $E(\bar{X}) = a_1$ and $E(\varepsilon_2) = 0$, the expected value of the MCSim estimator is

$$E\left(\tilde{F}_{MCSim}(x_0)\right) = a_2 + b_2 a_1 = \bar{F}(x_0)$$

i.e., the MCSim estimator is unbiased for every $m$, and its expectation trivially converges to the target mean for $m \to \infty$ (consistency).

For the coverage, the central $1 - \alpha$ interval of the MCSim distribution has to contain a new observation with probability $1 - \alpha$, i.e.,

$$P\left(Q_{\tilde{F}_{MCSim}(x_0)|X_1}(\alpha/2) \leq F(x_0) \leq Q_{\tilde{F}_{MCSim}(x_0)|X_1}(1 - \alpha/2)\right) = 1 - \alpha$$

where $Q$ denotes the quantile function of the MCSim distribution given the data $X_1$, which is, thus, a random quantity that depends on the data only. Unfortunately, no closed-form solution exists for finite $m$, because the MCSim distribution is the sum of a t-distributed and a normally distributed term. We therefore investigate the limit $m \to \infty$. Since $\bar{X} \to a_1$ and $S \to \sigma_1$ almost surely, and $t_{m-1}$ converges in distribution to a standard normally distributed variable $z_t$, we obtain

$$\tilde{F}_{MCSim}(x_0) = a_2 + b_2 \bar{X} + b_2 t_{m-1} S \sqrt{1 + \frac{1}{m}} + \varepsilon_2 \to a_2 + b_2 a_1 + b_2 \sigma_1 z_t + \sigma_2 z_2$$

$$= a_2 + b_2 a_1 + \sqrt{b_2^2 \sigma_1^2 + \sigma_2^2}\, z$$

with a standard normally distributed variable $z$ (the variables $z_t$ and $z_2$ are independent). The quantile function of this limit distribution, which no longer depends on the data, is

$$Q^{\infty}_{\tilde{F}_{MCSim}(x_0)}(\alpha) = a_2 + b_2 a_1 + \sqrt{b_2^2 \sigma_1^2 + \sigma_2^2}\, Q_z(\alpha)$$

The asymptotic coverage is therefore

$$P\left(Q^{\infty}_{\tilde{F}_{MCSim}(x_0)}(\alpha/2) \le F(x_0) \le Q^{\infty}_{\tilde{F}_{MCSim}(x_0)}(1-\alpha/2)\right)$$

$$= P\left(Q_z(\alpha/2) \le \sqrt{\frac{\sigma_2^2}{b_2^2\sigma_1^2 + \sigma_2^2}}\, z_2 \le Q_z(1-\alpha/2)\right)$$

Hence, the central quantity is the standard normally distributed variable $z_2$ scaled by the factor

$$\sqrt{\frac{\sigma_2^2}{\sigma_2^2 + b_2^2\sigma_1^2}}.$$

The asymptotic coverage equals $1-\alpha$ if $b_2^2\sigma_1^2 = 0$, and it is approximately $1-\alpha$ if $\sigma_2^2 \gg b_2^2\sigma_1^2$, because this factor is then (approximately) one. For $b_2^2\sigma_1^2 > 0$ the factor is smaller than one and the coverage is larger than the nominal level, i.e., it is overestimated; for $\sigma_2^2 = 0$ the coverage would be 100%. In the limit, fortunately, the coverage of the MCSim interval is therefore never below the target. However, how strongly it is overestimated depends on the unknown quantities $\sigma_1, \sigma_2$ and $b_2$, i.e. the extent is not known to the engineers in a modelling scenario. Since this is an asymptotic result, we see that even with an infinite amount of data the MCSim approach does not perform as it is expected from a well calibrated statistical model.

The same holds for the width of the interval. The central $1-\alpha$ interval of the distribution of the target random variable $F(x_0)$ is

$$\left[Q_{F(x_0)}(\alpha/2), Q_{F(x_0)}(1-\alpha/2)\right]$$

and its width is

$$Q_{F(x_0)}(1-\alpha/2) - Q_{F(x_0)}(\alpha/2) = \sigma_2\left(Q_z(1-\alpha/2) - Q_z(\alpha/2)\right)$$

In the limit, the average width of the MCSim interval is

$$E\left(Q^{\infty}_{\tilde{F}_{MCSim}(x_0)}(1-\alpha/2) - Q^{\infty}_{\tilde{F}_{MCSim}(x_0)}(\alpha/2)\right)$$

$$= \sqrt{b_2^2\sigma_1^2 + \sigma_2^2}\left(Q_z(1-\alpha/2) - Q_z(\alpha/2)\right)$$

Hence, the ratio of the average width of the MCSim interval to the true average width is

$$\sqrt{\frac{b_2^2\sigma_1^2 + \sigma_2^2}{\sigma_2^2}}$$

This ratio is one if $b_2^2\sigma_1^2 = 0$ and approximately one if $\sigma_2^2 \gg b_2^2\sigma_1^2$, but it is larger than one for $b_2^2\sigma_1^2 > 0$: the average width of the MCSim interval is then overestimated by this factor. As for the coverage, the amount of the overestimation depends on the unknown quantities $\sigma_1, \sigma_2$ and $b_2$, and again this result implies that collecting more data does not bring convergence to proper behavior.

As an even stronger counter example, the second unit operation is chosen to be a quadratic function with intercept, i.e. $f_2(x_1) = a_2 + c_2x_1^2 + \varepsilon_2$ with $\varepsilon_2 \sim N(0, \sigma_2)$ with $\sigma_2 > 0$ and $c_2 \neq 0$, which coefficients are known as well, i.e. $\tilde{f}_2 = f_2$. Given the true intercept model of the first UO and the true quadratic polynomial of the second UO, the true holistic mean function is $\bar{F}(x_0) = \bar{f}_2\left(\bar{f}_1(x_0)\right) = a_2 + c_2 a_1^2$. Again, analytically applying the MCSim approach, we obtain

$$\tilde{F}_{MCSim}(x_0) = \tilde{f}_2\left(\tilde{f}_1(x_0)\right)$$

$$= a_2 + c_2\bar{X}^2 + 2c_2\bar{X}t_{m-1}S\sqrt{1+\frac{1}{m}} + c_2 t_{m-1}^2 S^2\left(1+\frac{1}{m}\right) + \varepsilon_2.$$

An interesting observation is that $t_{m-1}^2$ is $F$-distributed with 1 and $m-1$ degrees of freedom, i.e., $\tilde{F}_{MCSim}(x_0)$ has no finite mean for $m \le 3$ and no finite variance for $m \le 5$. That means that it is possible that even though the intercept estimator has finite moments, the MCSim approach might not. Taking expectations gives

$$E\left(\tilde{F}_{MCSim}(x_0)\right) = a_2 + c_2 a_1^2 + c_2\frac{\sigma_1^2}{m} + c_2\sigma_1^2\left(\frac{m-1}{m-3}\right)\left(1+\frac{1}{m}\right)$$

$$= \bar{F}(x_0) + c_2\sigma_1^2 k_m,$$

with $\kappa_m \coloneqq \frac{m^2+m-4}{m(m-3)} > 1$ for all $m \ge 4$. Above, we used that

$$E(t_{m-1}^2) = \frac{m-1}{m-3},$$

and

$$E(\bar{X}^2) = a_1^2 + \frac{\sigma_1^2}{m},$$

where $z$ is standard normally distributed. Therefore, we have a bias of the MCSim mean prediction relative to the true holistic mean

$$E\left(\tilde{F}_{MCSim}(x_0)\right) - \bar{F}(x_0) = c_2\sigma_1^2 k_m$$

which depends on the variance of measurement error and quadratic coefficient. Its sign follows the sign of $c_2$, i.e. the direction of the bias is unknown to the engineers in reality. The finite sample size impact via $\kappa_m$ is great when $m$ is small, for example $\kappa_4 = 4$ or even still $\kappa_{15} \approx 1.3$, which is oftentimes the case in holistic process modelling. But also for $m \to \infty$, the bias $c_2\sigma_1^2\kappa_m \longrightarrow c_2\sigma_1^2$ does not vanish, i.e. the problem cannot be fixed with more data.

The coverage and width behavior cannot be investigated analytically nor asymptotically anymore, because the quantiles of $F_{MCSim}(x_0)$ are not accessible. These properties are evaluated in the simulation study in Section 3.2.

The linear example shows that MCSim can preserve the mean while still overestimating asymptotic coverage and interval width. The quadratic example shows an additional failure: nonlinear propagation introduces a mean bias that persists as the sample size increases. Table 1 summarizes the results established analytically.

**TABLE 1: SUMMARY OF EFFECTS DUE TO CHAINING WITH MCSIM APPROACH**

| Known second-stage model | Chaining method | Unbiased point predictor | Consistent point predictor | Coverage as sample size increases | Width relative to the true interval |
|---|---|---|---|---|---|
| **Affine** | MCSim | Yes | Yes (almost surely) | At or above the nominal level in the limit | Ratio at least 1 in the limit |
| Quadratic | MCSim | No | No | Not derived here; evaluated by simulation | Not derived here; evaluated by simulation |

We note that both examples are highly idealized, because the second unit operation is assumed to be known. If we would need to infer these coefficients as well or for even more consecutive UOs than two, the situation is expected to synergistically get worse. These scenarios are addressed by Monte Carlo simulation of the entire experiment, where this is the appropriate tool.

## 3.2 Simulation Studies

To investigate post-inference chaining beyond the analytical two-stage examples, we simulated a deterministic eight-stage chain and evaluated its outputs after stages 4 and 8. These endpoint models are denoted LQLQ and LQLQLQLQ, respectively, where L and Q indicate linear and quadratic stage means. The four-stage model is the prefix of the same eight-stage chain, not a separately specified data-generating process. The true stage means are:

$$\bar{f}_1(z) = z$$

$$\bar{f}_2(z) = 0.5z + 0.1z^2$$

$$\bar{f}_3(z) = 5 - z$$

$$\bar{f}_4(z) = \frac{20}{7}z - \frac{20}{49}z^2$$

$$\bar{f}_5(z) = z$$

$$\bar{f}_6(z) = 5 - 0.6z - 0.08z^2$$

$$\bar{f}_7(z) = 5 - z$$

$$\bar{f}_8(z) = \frac{200}{49} + \frac{60}{49}z - \frac{20}{49}z^2$$

Each stage maps [0, 5] onto [0, 5]. The linear stages alternate between increasing and decreasing functions. The quadratic stages alternate between monotone functions (stages 2 and 6) and functions with an interior maximum (stage 4 at z = 3.5; stage 8 at z = 1.5). The latent output at stage $i$ is $\bar{F}_i(x) = \bar{f}_i(\bar{F}_{i-1}(x))$, with $\bar{F}_0(x) = x$. Only latent stage means enter the next true stage; measurement errors are added to the measured outputs, not propagated through the true chain.

For each sample size $m \in \{7,15,30\}$, we performed $R = 1000$ independent outer repetitions. The $m$ initial inputs were equally spaced over [0, 5], including both endpoints. Each run $k$ provided one measured output at each of the eight stages:

$$X_i^k = \bar{F}_i(x_0^k) + \varepsilon_i^k, \ \ \varepsilon_i^k \sim N(0,\sigma), \ \ \sigma = 0.5,$$

where the measurement errors were independent across runs and stages, with the same standard deviation at every stage. Thus, the measurement-error standard deviation was 10% of the stage signal range of 5. Predictions were evaluated on the grid $G = \{0,0.25,\dots,5\}$. The true endpoint mean on this grid is $\mu_i(g) = \bar{F}_i(g)$, for $i = 4$ or 8 and $g \in G$.

Each stage was fitted separately by ordinary least squares, using an intercept and a linear term for L stages, and an intercept, linear term and squared input for Q stages. Stage 1 used the known initial inputs. In the observed-to-observed variant shown in the main figures, each later stage regressed its measured output on the measured output of the preceding stage. The latent-to-observed variant instead used the true preceding latent outputs as training inputs. This latter variant is an oracle diagnostic, not a feasible procedure with unknown latent values: it removes measurement error from the training predictors while retaining the same MCSim propagation rule. It therefore distinguishes an errors-in-variables effect from the effect of forwarding predictive draws after inference. Both variants used the same generated training data within each outer repetition.

The endpoint prediction interval was the empirical 5th to 95th percentile interval of the $B$ endpoint draws (nominal level 90%, $\alpha = 0.1$). No predictive draw was truncated to $[0,5]$; consequently, propagated inputs could leave the training range, where the fitted polynomials and leverage terms were still evaluated.

As we already saw in Section 3.1 the mean in the MCSim approach might not exist. Thus, the code tracks a predictive tail index $\tau_i$. The predictive mean exists only when the tail index exceeds 1; a finite variance requires an index exceeding 2. Thus, a finite $B$ always yields a numerical sample mean, but that value need not estimate an existing population mean. Even when the mean exists, small tail indices can make it unstable.

For each endpoint and outer repetition $r$, the code selected the mean of the $B$ draws if the tail index exceeded 1; otherwise it selected their median. Denote this selected statistic by $\hat{\mu}_{i,r}(g)$. Its relative deviation from the true endpoint mean was

$$\delta_{i,r}(g) = \frac{\hat{\mu}_{i,r}(g) - \mu_i(g)}{\Delta_i}, \ \ \Delta_i = max\mu_i(g) - min\mu_i(g), \ \ g \in G.$$

The denominator is the range of the true endpoint means over the evaluation grid, not automatically 5.

Coverage concerns one new measured endpoint observation at a fixed grid point. The reported coverage is the mean over the 1000 outer repetitions. Interval width was likewise averaged over repetitions after division by the true central 90% normal-interval width. A width ratio of 1 corresponds to the interval width with known endpoint mean and measurement-error standard deviation. It is an oracle reference, not a finite-sample optimum for intervals based on fitted models. For the summary curves, pointwise 90% percentile bootstrap confidence intervals were calculated from 2.000 resamples of the outer repetitions.

Figure 1 shows that MCSim point predictions deviate substantially from the true endpoint means. For LQLQ, relative mean bias ranges from −22.1% to −16.9% at $m = 7$, from −13.0% to +18.8% at $m = 15$ and from −11.5% to +29.3% at $m = 30$ over the grid. A larger training set therefore does not uniformly remove the discrepancy. For LQLQLQLQ, the median-based relative deviations range from −44.8% to +41.4% at $m = 7$ and from −38.2% to +39.5% at $m = 15$. At $m = 30$, where the predictive mean exists, relative mean bias ranges from −45.0% to +34.1%. We must note that wherever the mean prediction does not exist anymore, the MCSim approach is not applicable anymore. Consequently, their ordering must not be interpreted as a monotone sample-size effect on mean bias. The pronounced input-dependent deviations demonstrate that the distortion is not merely a common constant offset.

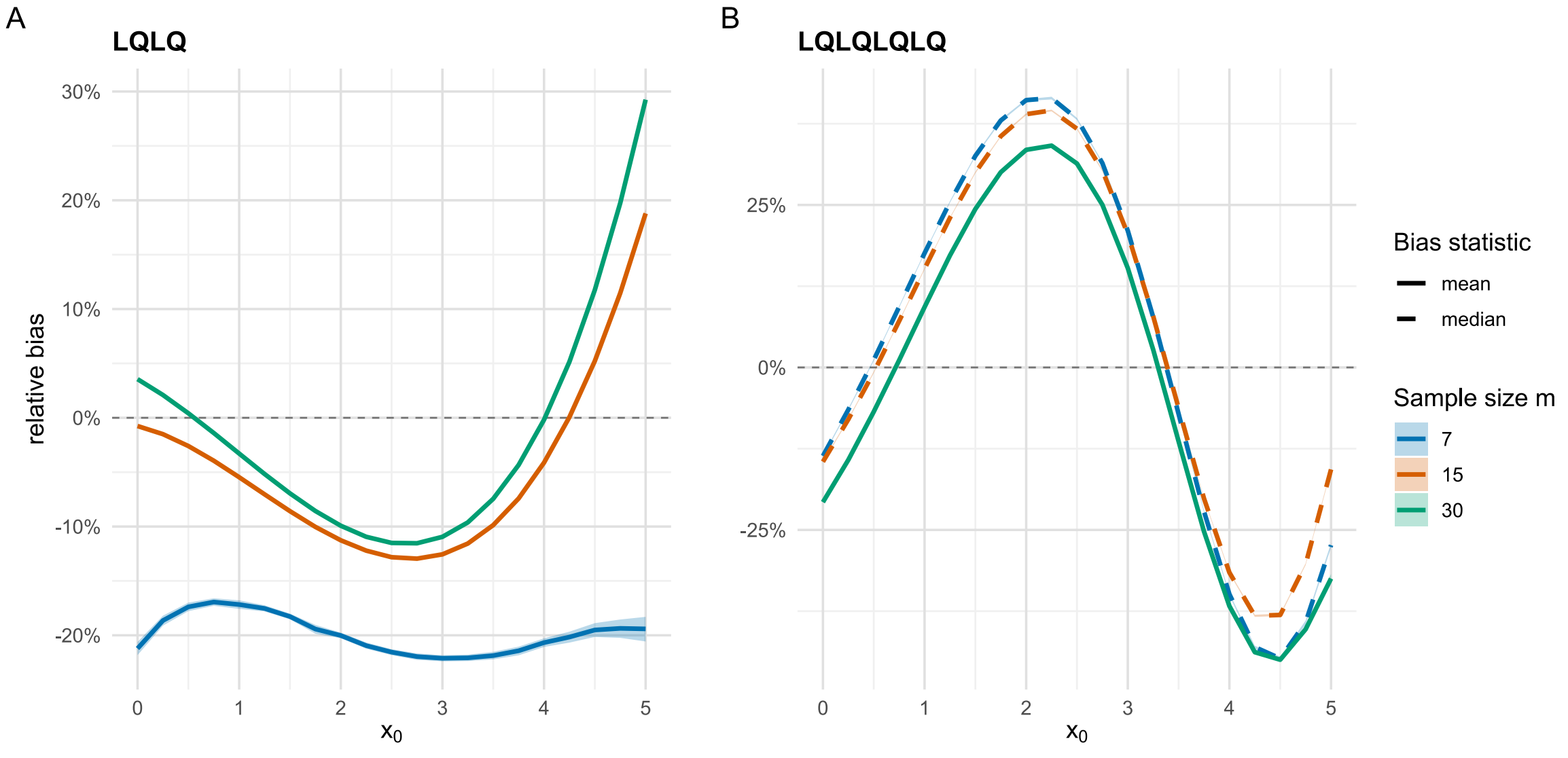

**Figure 1: Relative endpoint deviations from the true mean under observed-to-observed MCSim training. Panels A and B show LQLQ (four stages) and LQLQLQLQ (eight stages). Blue, orange and green correspond to m = 7, 15 and 30. Values are normalized by the true endpoint-mean range over the 21-point input grid. Mean-based cells report the mean across 5,000 repetitions of the predictive-mean deviation. For stage 8 at m = 7 and 15, the predictive mean does not exist; the statistic is instead the median across repetitions of the predictive-median deviation. Ribbons are pointwise 90% outer-repetition bootstrap confidence intervals (2,000 resamples); B = 5,000 predictive chains per grid point.**

Reliable prediction also requires appropriate coverage for a new measured endpoint observation. Figure 2 shows that the observed-to-observed MCSim intervals exceed the nominal 90% target at every evaluated grid point by far. Coverage ranges from 95.4% to 100% for LQLQ and from 94.1% to 100% for LQLQLQLQ across the three sample sizes. This is conservative behavior in these particular simulations, not evidence of well-calibrated 90% intervals. Increasing $m$ does not bring the curves uniformly towards the nominal target.

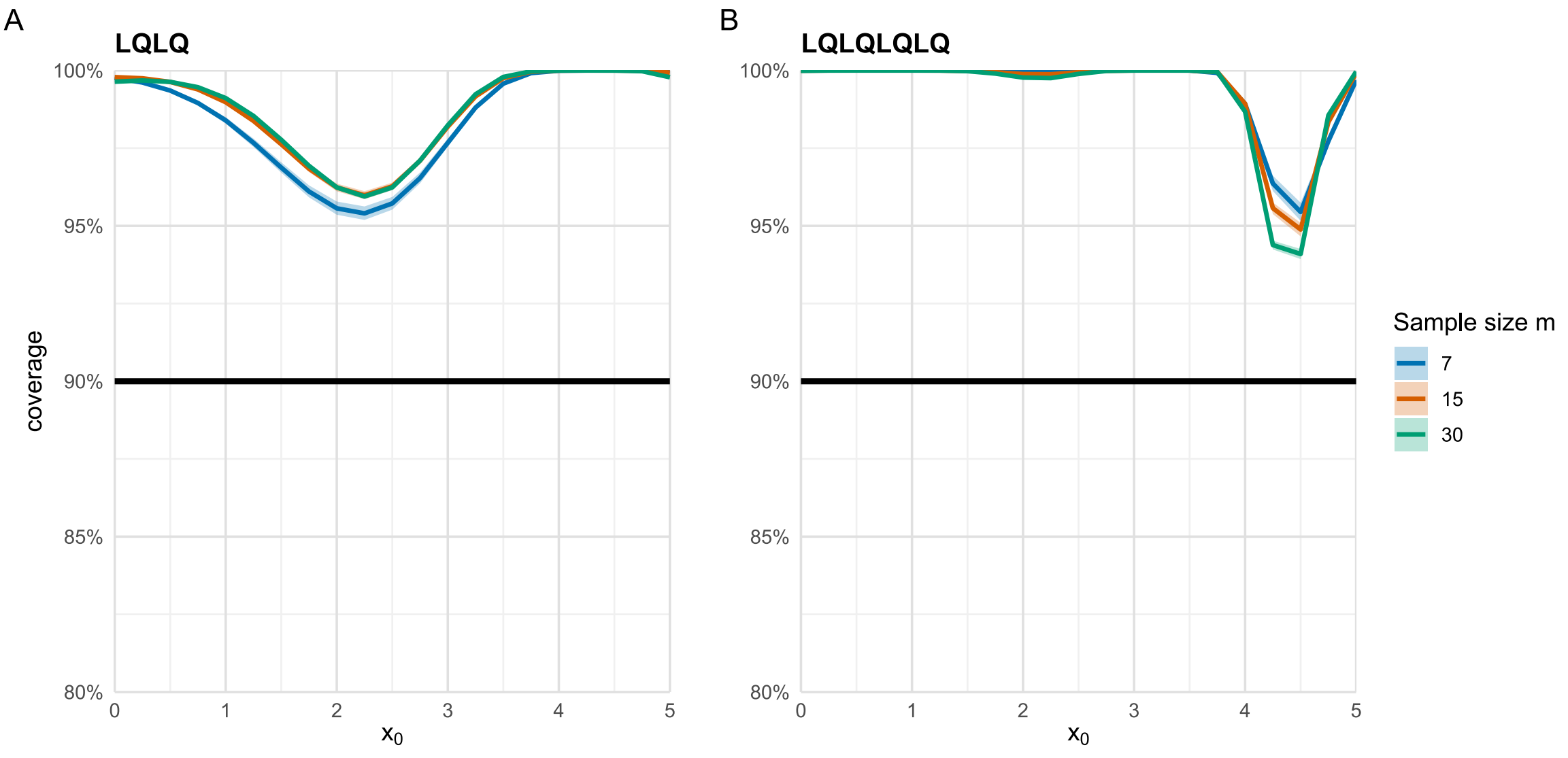

**Figure 2: Pointwise coverage of nominal 90% MCSim prediction intervals for a new measured endpoint observation, using observed-to-observed training. Panels A and B show LQLQ and LQLQLQLQ; colors identify m = 7, 15 and 30 as in Figure 1. Coverage is the true normal probability integrated over each fitted interval and then averaged over 5,000 outer repetitions. The black horizontal line is the 90% target. Ribbons are pointwise 90% outer-repetition bootstrap confidence intervals. B = 5,000 predictive chains per grid point.**

Figure 3 quantifies the loss of precision behind this overcoverage. For LQLQ, average interval widths are 3.66–11.16 times the true normal-interval width at $m = 7$, 2.19–6.30 times at $m = 15$ and 1.87–4.97 times at $m = 30$. For LQLQLQLQ, the corresponding ranges are 10.26–656.78, 4.83–12.93 and 4.11–4.97. Thus, even m = 30 leaves endpoint intervals approximately four to five times the oracle width for the eight-stage chain. The especially large values for $m = 7$ occur in a regime of severe tail amplification. Widths substantially above the oracle reference are not by themselves proof of an invalid finite-sample interval, but the combination of very large widths, systematic point-prediction deviations and coverage far above the requested level makes this implementation poorly calibrated and imprecise in the investigated designs.

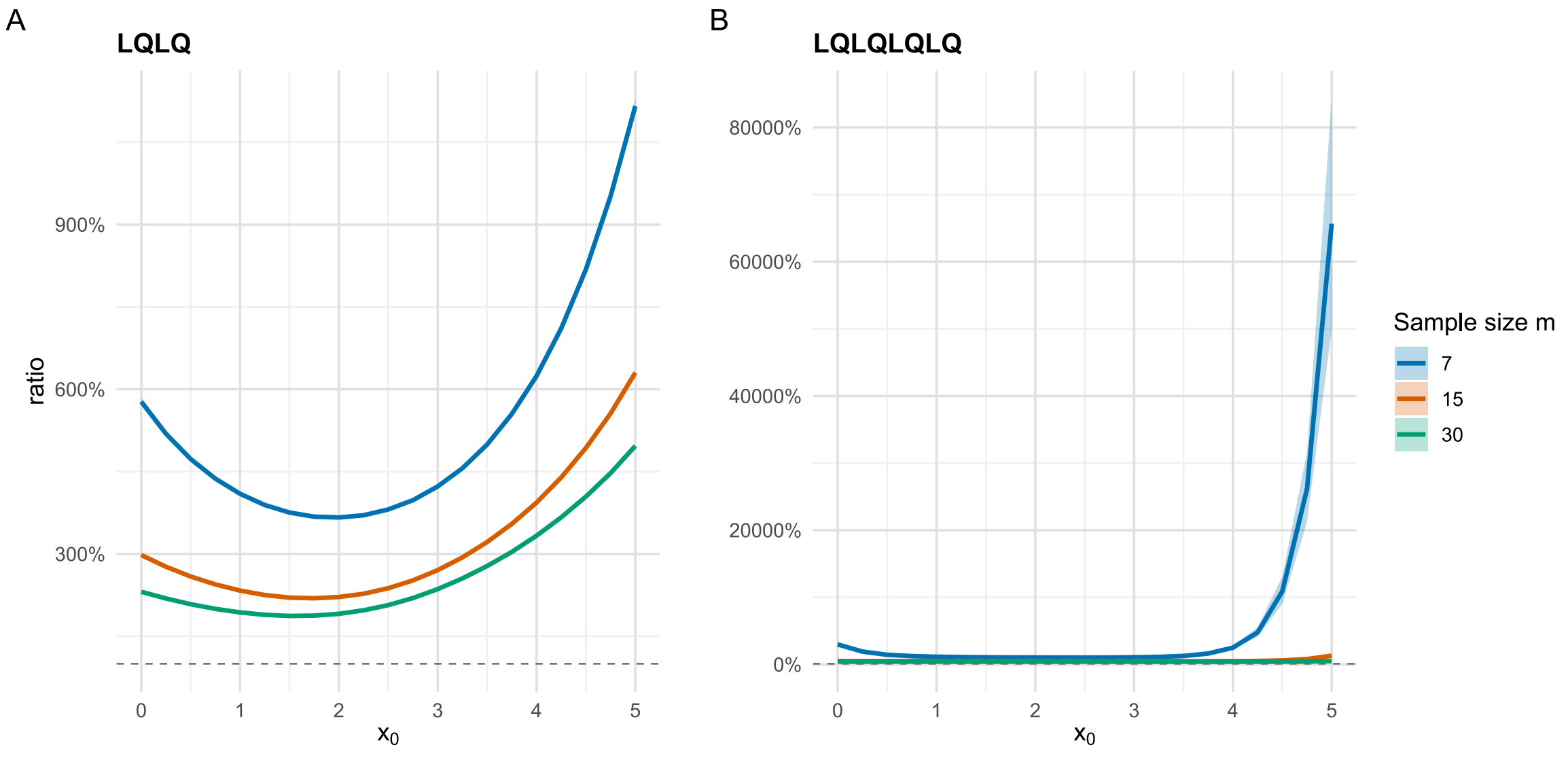

FIGURE 3: MEAN MCSIM INTERVAL-WIDTH RATIO UNDER OBSERVED-TO-OBSERVED TRAINING. PANELS A AND B SHOW LQLQ AND LQLQLQLQ; COLORS IDENTIFY SAMPLE SIZES AS IN FIGURE 1. EACH 90% PREDICTION-INTERVAL WIDTH IS DIVIDED BY THE TRUE CENTRAL 90% NORMAL-INTERVAL WIDTH AND AVERAGED OVER 1000 OUTER REPETITIONS. THE PERCENTAGE SCALE EXPRESSES A RATIO OF 1 AS 100%, WHICH IS THE DASHED ORACLE REFERENCE. RIBBONS ARE POINTWISE 90% OUTER-REPETITION BOOTSTRAP CONFIDENCE INTERVALS; B = 5,000 PREDICTIVE CHAINS PER GRID POINT.

The latent-to-observed diagnostic (Figure S1, Figure S2 and Figure S3) separates the input-measurement-error issue from post-inference propagation. In that variant, every stage is fitted using the true preceding latent output, so the training predictors are measurement-error free. Nevertheless, for LQLQ at $m = 30$, relative mean bias still ranges from −14.5% to −5.9%, and mean interval widths range from 1.42 to 7.03 times the oracle width. For LQLQLQLQ at $m = 30$, widths range from 5.56 to 27.83 times the reference and coverage falls to about 83.1% at some inputs despite being near 100% elsewhere. These results show that correcting the training-input measurement error alone does not repair the forwarding of full predictive draws. They also caution against extending the analytical linear two-stage overcoverage result to arbitrary nonlinear multi-stage chains. The simulation supports the need for a different post-inference construction, evaluated next using measurement-error setting and models.

# 4 Holistic Retraining

The main hurdles in holistic process modelling are the error-in-variable problem and the missed interdependencies due to separate training as discussed in more detail in the introduction. In this section we propose an algorithm that aims to holistically retrain the single models to reconnect them to a holistic model that has drastically reduced bias, fulfills the target coverage approximately and has a smaller interval width in comparison with the MCSim approach. To test this algorithm, we apply it to the two chains introduced in the previous section as well as additionally to a bivariate 6-stage model shown in the supplement.

## 4.1 Model-based Bootstrap

The algorithm consists of two phases: the first constructs the ***initial*** holistic or multi-stage model that is used for the holistic mean prediction and the second phase generates a set of ***bootstrap*** holistic models which are used to compute the uncertainty of the holistic prediction.

The core of the algorithm is a model-based bootstrap, i.e., we train models on the data and reuse them as data-generating processes. This already highlights the core requirement for the algorithm: the model type needs to know its own estimated error distribution. This is usually given for (mixed effect) linear parameter models or whenever the single models can be represented by a bootstrap model (parametric, residual or other bootstraps) the algorithm is applicable.

The two phases of the algorithm to train the holistic model can be found in Pseudo Code 1 and 2 and the algorithm to compute its predictions with uncertainty in Pseudo Code 3. Its variables are explained in Table 2.

**TABLE 2: VARIABLES USED IN THE PSEUDO CODES**

| Variable | Explanation |
|---|---|
| $n$ | number of stage-models in the multi-stage model |
| $i$ | index to (initial or bootstrap) stage-model and mean predictions ($i \in \{1, \dots, n\}$) as well as measured data ($i \in \{0, 1, \dots, n\}$) |
| $D_i$ | output data from stage $i$ and input data to stage $i + 1$ |
| $M_i$ | $i$-th initial stage-model |
| $M_i(x, "m")$ | generating **m**ean predictions from $i$-th initial stage-model for input data $x$ |
| $M_i(x, "s")$ | drawing **s**amples from the $i$-th initial stage-model error distribution for input data $x$ |
| $\overline{D}_i$ | mean predictions from $i$-th initial stage-model |
| $B$ | number of bootstrap loops and bootstrap (multi-)stage models |
| $b \in \{1, \dots, B\}$ | index to bootstrap iteration and bootstrap (multi-)stage model |
| $D_i^b$ | bootstrap output data from stage $i$ of the $b$-th bootstrap loop |
| $M_i^b$ | $i$-th bootstrap stage-model of the $b$-th bootstrap loop |
| $M_i^b(x, "m")$ | generating **m**ean predictions from $i$-th bootstrap stage-model of the $b$-th bootstrap loop for input data $x$ |
| $M_i^b(x, "s")$ | drawing **s**amples from the $i$-th bootstrap stage-model uncertainty distribution of the $b$-th bootstrap loop for input data $x$ |
| $\overline{D}_i^b$ | mean predictions from $i$-th bootstrap stage-model of the $b$-th bootstrap loop |

**PSEUDO CODE 1:**

For $i$ in 1 till $N$:
    If $i = 1$
        Train $M_1$ on $(D_0, D_1)$
        Set $\overline{D}_1$ to $M_1(D_0, "m")$
    Else
        Train $M_i$ on $(\overline{D}_{i-1}, D_i)$
        Set $\overline{D}_i$ to $M_i(\overline{D}_{i-1}, "m")$

### PSEUDO CODE 2:

For $b$ in 1 till $B$:
    For $i$ in 1 till $N$:
        If $i = 1$
            Set $D_1^b$ to $M_1(D_0, "s")$
            Train $M_1^b$ on $\left(D_0, D_1^b\right)$
            Set $\overline{D}_1^b$ to $M_1^b(D_0, "m")$
        Else:
            Set $D_i^b$ to $M_i(\overline{D}_{i-1}, "s")$
            Train $M_i^b$ on $\left(\overline{D}_{i-1}^b, D_i^b\right)$
            Set $\overline{D}_i^b$ to $M_i^b(\overline{D}_{i-1}^b, "m")$

In prose, the algorithm reads: in the first phase, we build an initial multi-stage model by training the first stage model $M_1$ on the input data $D_0$ (that is assumed to be deterministic) and output data $D_1$. In the next step, the second model is not trained on $D_1$ as input data, but on $\overline{D}_1$, i.e. the mean predictions from the first stage model $M_1$ applied to $D_0$. With this, we approximately mitigate the error-in-variable problem, because the mean prediction excludes the measurement error as best as possible. This gets better, the more data is available, i.e. the more precise the estimation is. This is now repeated until the last stage.

This initial holistic model is now used to establish multi-stage bootstrap uncertainty in the second training phase. For each bootstrap loop iteration $b$, we apply the first initial stage model $M_1$ on $D_0$ as inputs to sample a bootstrap data set $D_1^b$ that has the same dimensions as the original data set $D_1$ and train a bootstrap model $M_1^b$ on $D_0$ and $D_1^b$. This model is used to provide a mean prediction $\overline{D}_1^b$ and, in the second stage, model $M_2$ is used to generate a bootstrap data set $D_2^b$ with $\overline{D}_1$ as inputs. On this generated pair $\left(\overline{D}_1^b, D_2^b\right)$ we then train a second stage bootstrap model $M_2^b$. This is now repeated until the last stage and for all bootstrap iterations. Here, we used the same strategy as in the training of the initial multi-stage model to approximately eliminate the error-in-variable problem. Another reason for the strategy of forwarding the bootstrap mean predictions is that the bootstrap data dependency is carried through the whole chain re-introducing the dependencies between model coefficients, i.e., the second problem of holistic process modelling is addressed.

The next step is the prediction of a future state by the holistic process model. This is described in Pseudo Code 3 which also uses the notations of Table 2. The algorithm is again divided into two phases. Given a future first stage input $D_0$, the first phase computes the mean prediction by simply executing the initial model chain. The second phase evaluates all bootstrap multi-stage models at $D_0$ and approximates any desired statistical property from the obtained bootstrap samples in whichever way seems suitable. In our case we used a studentized prediction root [20].

### PSEUDO CODE 3:

    For $i$ in 1 till $N$:
        If $i = 1$
            Set $\overline{D}_1$ to $M_1(D_0, "m")$
        Else
            Set $\overline{D}_i$ to $M_i(\overline{D}_{i-1}, "m")$
    For $b$ in 1 till $B$:
        For $i$ in 1 till $N$:
            If $i = 1$

Set $\overline{D}_1^b$ to $M_1^b(D_0, "m")$

Set $D_1^b$ to $M_1^b(D_0, "s")$

Else:

Set $\overline{D}_i^b$ to $M_i^b(\overline{D}_{i-1}^b, "m")$

Set $D_i^b$ to $M_i^b(\overline{D}_{i-1}^b, "s")$

It is noteworthy that the mean predictions with statistical uncertainty intervals cannot only be computed at the last stage $n$ of the holistic process, but also at any other intermediate stage $i \in \{2, \dots, n-1\}$. The next subsection investigates the just presented algorithm on the LQLQ and LQLQLQLQ models.

## 4.2 Simulation Studies applying the model-based Bootstrap

We evaluated the holistic bootstrap on the LQLQ and LQLQLQLQ endpoints from Section 3.2, using the same true stage functions, measurement-error standard deviation 0.5, sample sizes $m = 7, 15$ and $30$. There were 1000 outer repetitions and B = 1000 refitted bootstrap chains per repetition. The point predictor was the noise-free initial fitted-to-observed composition, not an average of bootstrap predictions. Relative mean bias and interval-width ratios use the same true endpoint ranges and normal-interval reference as the MCSim study.

Figure 4 shows substantially smaller relative mean biases than the MCSim results. For LQLQ, the ranges over the grid are −1.34% to +1.11% at $m = 7$, −0.73% to +1.01% at $m = 15$ and −0.38% to +0.57% at $m = 30$. For LQLQLQLQ, they are −6.98% to +7.87%, −4.97% to +4.84% and −3.16% to +2.65%, respectively. The longer chain therefore retains visible, input-dependent bias, although its magnitude decreases over the investigated sample sizes. These finite-sample results are compatible with improved estimation as $m$ increases, but do not establish general consistency.

Figure 5 shows coverage much closer to the 90% target than the main MCSim study. For LQLQ, pointwise coverage ranges are 88.26–90.02%, 88.76–90.76% and 89.30–90.72% for $m = 7, 15$ and 30. For LQLQLQLQ, the corresponding ranges are 88.18–92.16%, 86.04–90.64% and 88.32–90.68%.

Figure 6 shows considerably less width inflation than MCSim. For LQLQ, mean width ratios range from 1.43 to 1.73 at $m = 7$, 1.14 to 1.38 at $m = 15$ and 1.06 to 1.23 at $m = 30$. For LQLQLQLQ, they range from 1.97 to 2.78, 1.36 to 1.65 and 1.16 to 1.42. Widths decrease towards the known-parameter normal reference as $m$ increases, and the longer chain generally requires wider intervals. A ratio above 1 is expected to some extent because the oracle reference excludes parameter-estimation uncertainty; it does not alone imply conservatism. Indeed, Figure 5 contains local undercoverage despite ratios above 1. At $m = 30$, the eight-stage bootstrap ratios of 1.16–1.42 contrast with 4.11–4.97 for observed-to-observed MCSim, while achieving coverage far closer to the requested level. This is evidence of a useful improvement within the tested design.

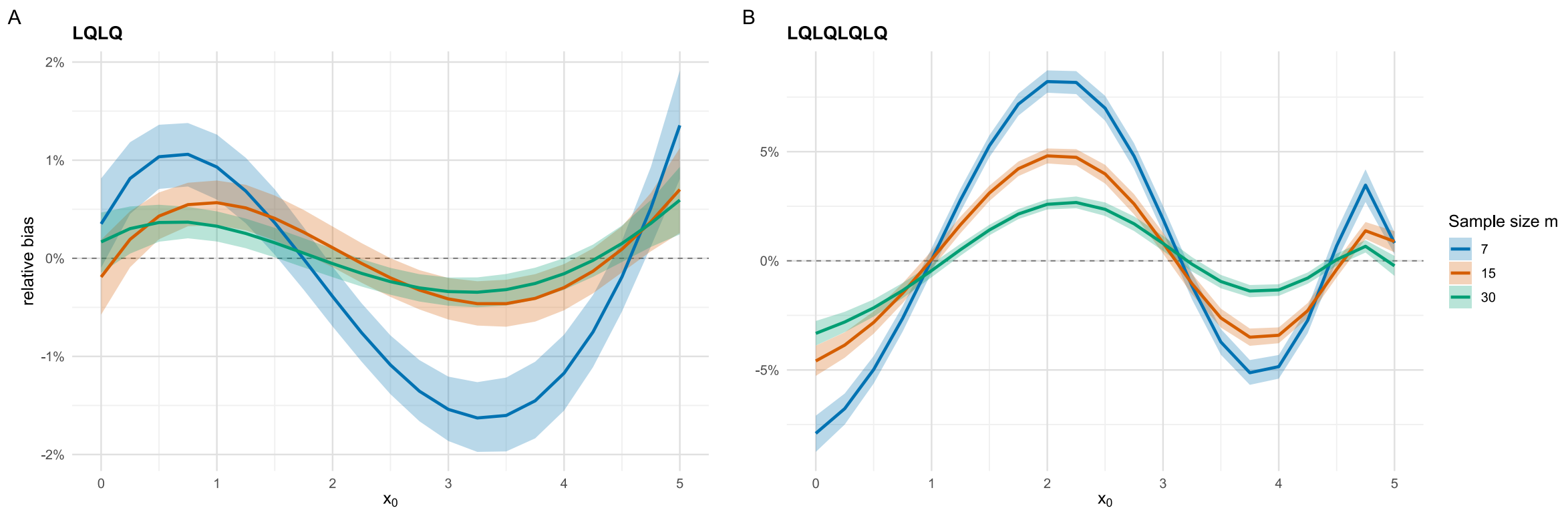

Figure 4: Relative mean bias of the initial fitted-to-observed chain used by the fixed-center studentized bootstrap. Panels A and B show LQLQ and LQLQLQLQ; blue, orange and green denote m = 7, 15 and 30. Bias is normalized by the true endpoint-mean range over the evaluation grid and averaged over 1000 outer repetitions. Ribbons are pointwise 90% percentile bootstrap confidence intervals from 2,000 resamples of the outer repetitions.

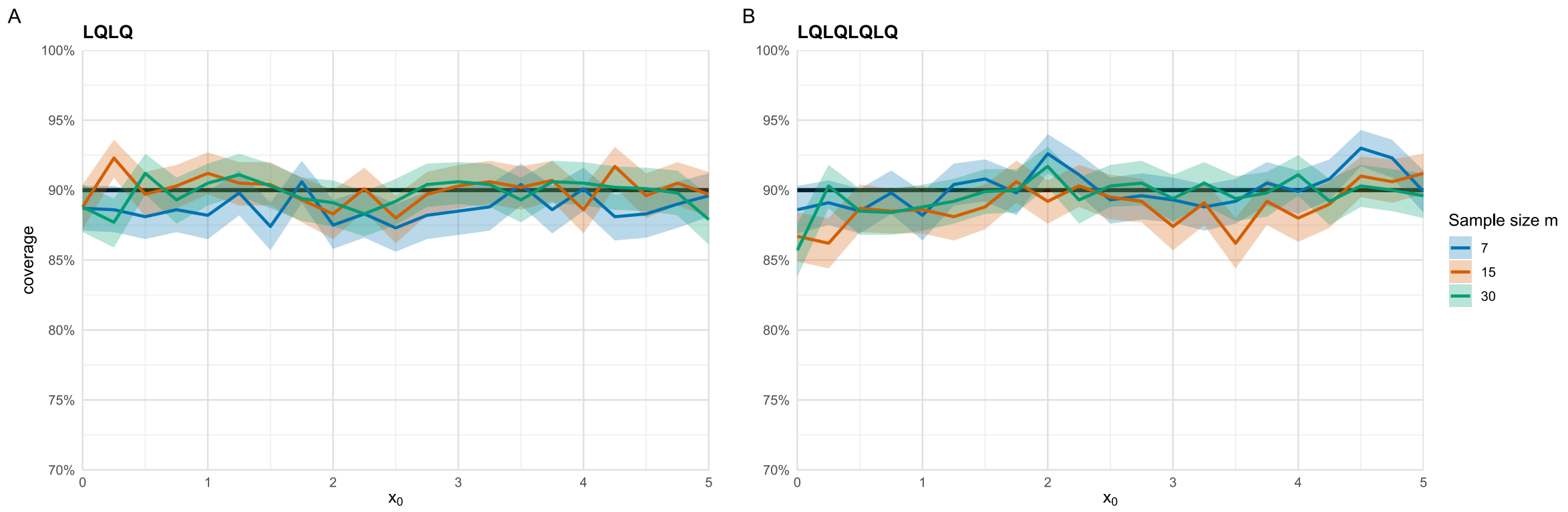

Figure 5: Pointwise coverage of nominal 90% fixed-center studentized bootstrap prediction intervals. Panels A and B show LQLQ and LQLQLQLQ; colors identify m = 7, 15 and 30 as in Figure 4. Coverage is the proportion of 1000 repetitions in which one fresh normal endpoint observation lies inside the fitted interval. The black line is the 90% target. Ribbons are pointwise 90% outer-repetition bootstrap confidence intervals; B = 5,000 refitted chains per repetition.

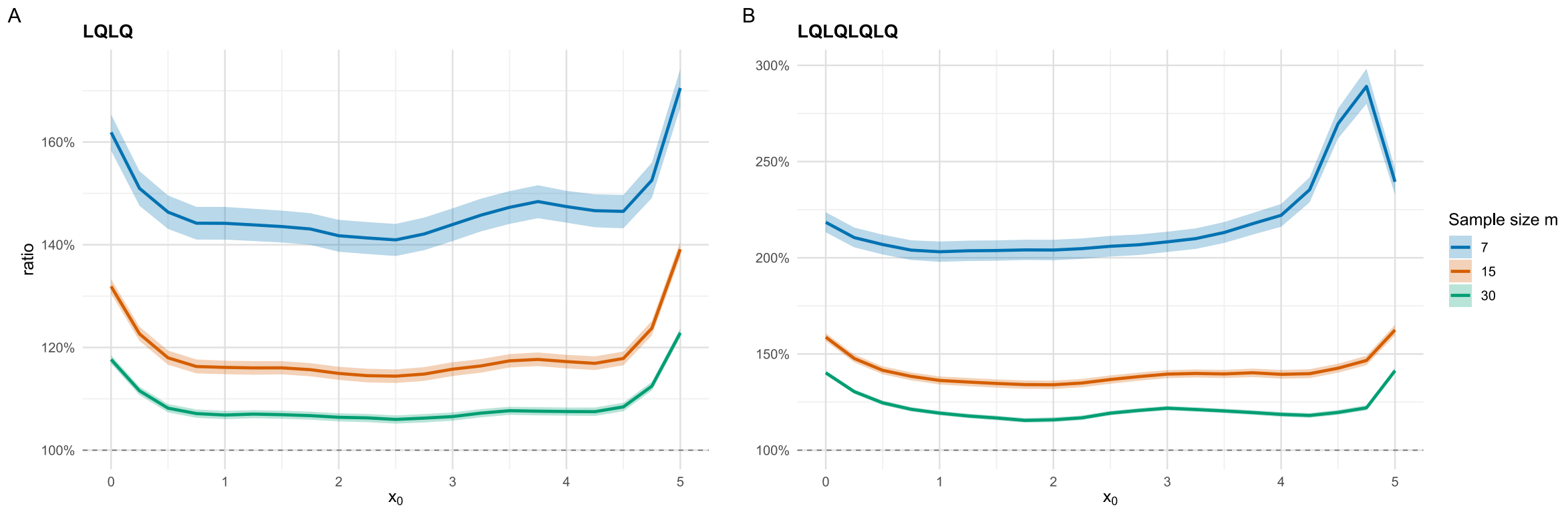

Figure 6: Mean width ratios of the fixed-center studentized bootstrap prediction intervals. Panels A and B show LQLQ and LQLQLQLQ; colors identify sample sizes as in

**Figure 4. Widths are divided by the true central 90% normal-interval width and averaged over 1000 outer repetitions. The dashed line is this known-parameter reference. Ribbons are pointwise 90% confidence intervals from 2,000 resamples of the outer repetitions; B = 5,000 bootstrap refits.**

A bivariate 6-stage illustration is retained in the supplement: Figures S4–S9 describe its bivariate stage functions, while Figures S9–S12 contain results from the holistic bootstrap. The example with $n = 60$ batches shows the same tendencies for the final output, but also for the intermediate output after UO2: minimal relative bias (below 15% even after UO6), coverage is approximately met and the interval width is close to 100% except at the edges of the design space (Figures S6-8). The latter is a wanted behavior since models tend to be less precise at the edges to extrapolation, i.e., it is also seen for UO1, where no chaining has happened yet.

# 5 Conclusion

Chaining separately fitted stage models by forwarding complete predictive draws can confuse uncertainty in the estimated latent stage means with measurement error in a future observed output. The analytical intercept-to-linear example shows finite but also asymptotic overcoverage and width inflation, even though its mean remains unbiased. Introducing a quadratic second stage adds a finite but also asymptotic mean bias. The four- and eight-stage simulations show that the resulting distortions can be substantial, input-dependent and not eliminated by larger training samples over the investigated range. In both investigations, a crucial observation was that the first as well as second (and at some point also higher) moments can get undefined. The latent-input diagnostic demonstrates that training-input measurement error alone does not explain the failure. It also shows local undercoverage, so the special-case linear overcoverage result must not be generalized to arbitrary nonlinear chains.

The holistic retraining via a model-based bootstrap provides a substantially better-performing alternative for these simulated chains. It first estimates a fitted-to-observed mean chain, then refits that estimator on stage-specific pseudodata centered at fixed initial fitted values. Relative mean bias in the four-stage chain is approximately within −1.34% to +1.11% at $m = 7$ and −0.38% to +0.57% at $m = 30$. For the eight-stage chain, the corresponding ranges are −6.98% to +7.87% and −3.16% to +2.65%. At $m = 30$, mean width ratios of 1.16–1.42 for the eight-stage bootstrap compare with 4.11–4.97 for observed-to-observed MCSim. These gains concern the combined change in point estimation and interval construction, not studentization in isolation.

The bootstrap is nevertheless not uniformly exact. Pointwise coverage across the evaluated scenarios ranges from 86.04% to 92.16% at a nominal 90%, with visible local undercoverage in the eight-stage chain. Widths remain above the known-parameter reference, especially for small samples and longer chains. Model-form misspecification, correlated or heteroscedastic errors, other designs and other model families have not been evaluated. Fitted-mean regressors remain estimated and can therefore mitigate rather than generally eliminate errors-in-variables effects.

Within the stated scope, the results support the holistic bootstrap as a promising basis for more informative endpoint prediction.

# Supplements

## S1. MCSim diagnostic with latent training inputs

This diagnostic uses the same simulated process, sample sizes, measured responses, prediction grid and MCSim propagation scheme as the main study. Only the training predictors change: later-stage models use the true preceding latent output rather than its measured value. The diagnostic therefore removes the errors-in-variables component from the training inputs without removing the predictive noise that MCSim subsequently propagates. Its purpose is not to propose an implementable oracle method, but to test whether training-input measurement error is the sole source of the observed distortions.

Figure S1 shows persistent discrepancies even with latent training inputs. For LQLQ, mean bias remains negative over the grid: −38.9% to −13.8% at $m = 7$, −19.9% to −7.5% at $m = 15$ and −14.5% to −5.9% at $m = 30$. For LQLQLQLQ at $m = 30$, mean bias reaches −378.9% near the right boundary.

Figure S2 shows LQLQ coverage of approximately 90.7–100%. The eight-stage diagnostic instead ranges from approximately 82.8% to 100% across sample sizes, with a localized trough near the upper end of the input range. Figure S3 shows that undercoverage can coexist with very large widths: at $m = 7$, eight-stage mean widths range from 11.11 to 1733.35 times the reference; even at $m = 30$ they range from 5.56 to 27.83 times. These are distinct grid-dependent features, not a claim that the maximum width occurs where coverage is minimal. Nonlinearity changes the location, shape and tails of the predictive distribution; wider intervals alone need not restore correct coverage. Taken together, these results rule out training-input measurement error as the sole explanation of the MCSim failure in these scenarios.

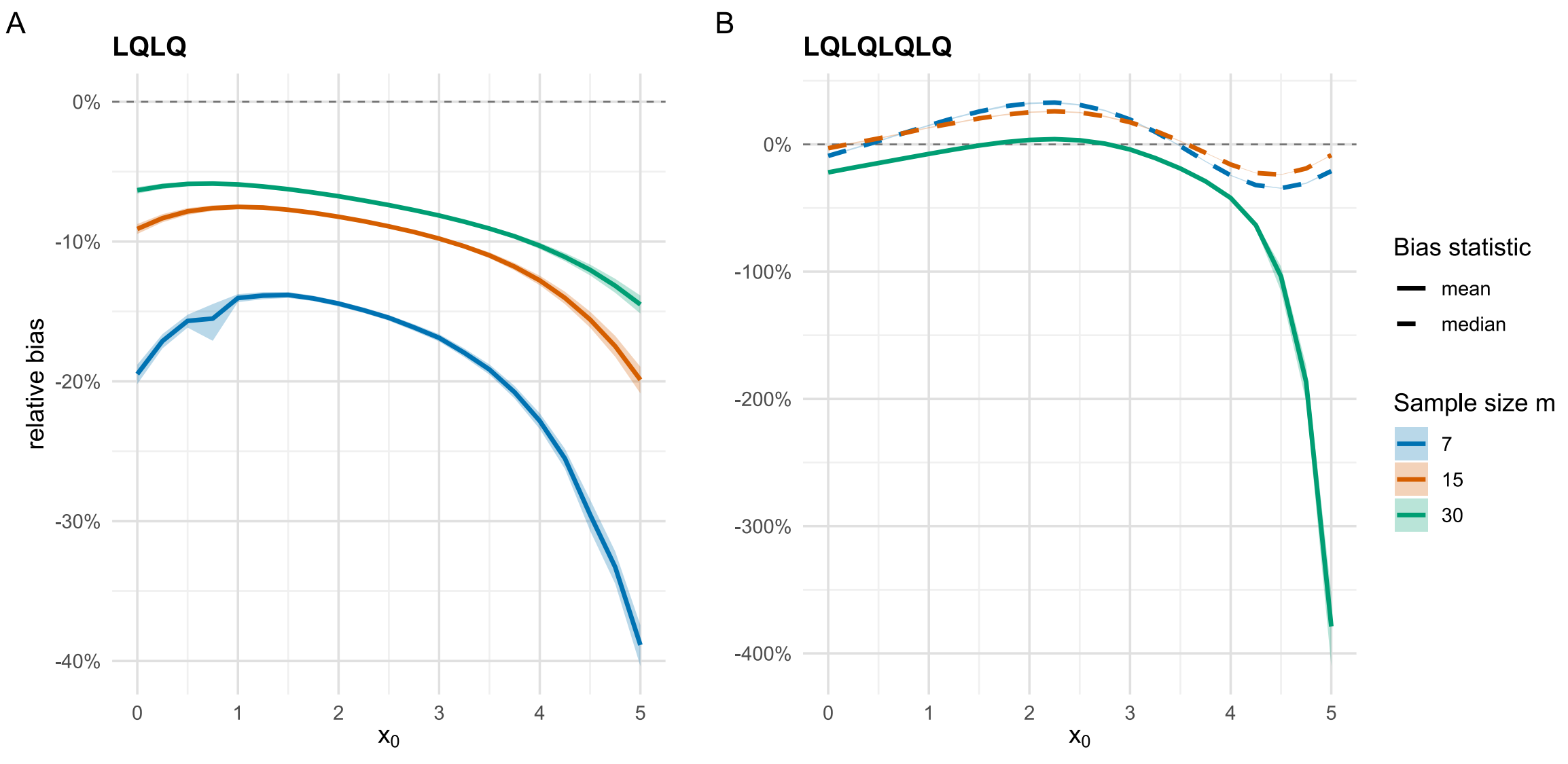

**FIGURE S1: RELATIVE ENDPOINT DEVIATIONS UNDER LATENT-TO-OBSERVED TRAINING. PANELS A AND B SHOW LQLQ AND LQLQLQLQ; COLORS DENOTE $m = 7, 15$ AND $30$. NORMALIZATION, AGGREGATION AND POINTWISE 90% BOOTSTRAP BANDS ARE AS IN FIGURE 1.**

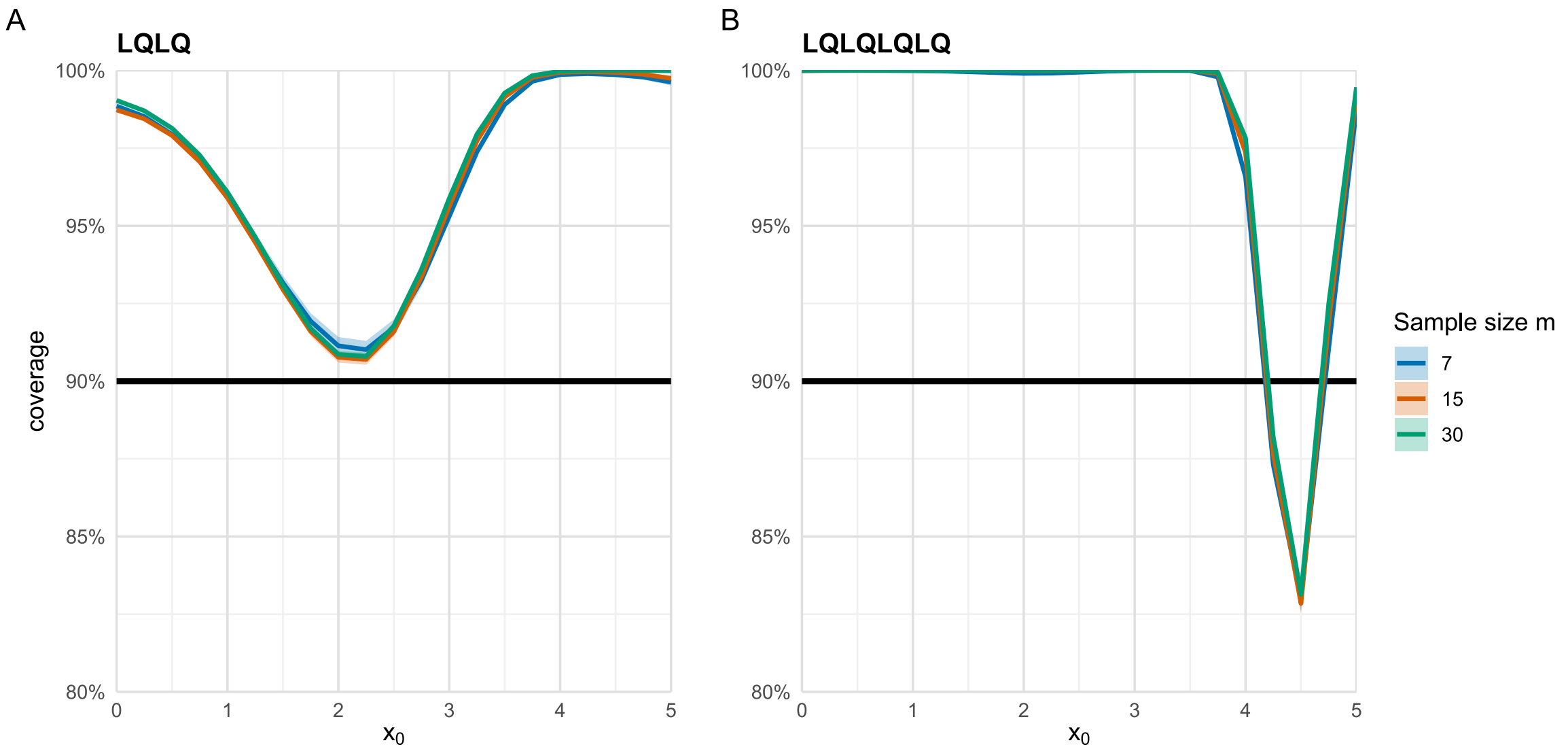

FIGURE S2: POINTWISE COVERAGE OF NOMINAL 90% MCSIM PREDICTION INTERVALS WITH LATENT-TO-OBSERVED TRAINING. PANELS A AND B SHOW LQLQ AND LQLQLQLQ; COLORS IDENTIFY $m = 7, 15$ AND $30$. THE BLACK LINE IS THE 90% TARGET. RIBBONS ARE POINTWISE 90% OUTER-REPETITION BOOTSTRAP CONFIDENCE INTERVALS. THE LOCALIZED UNDERCOVERAGE IN PANEL B SHOWS THAT THE OVERCOVERAGE OF THE MAIN OBSERVED-TO-OBSERVED STUDY IS NOT UNIVERSAL.

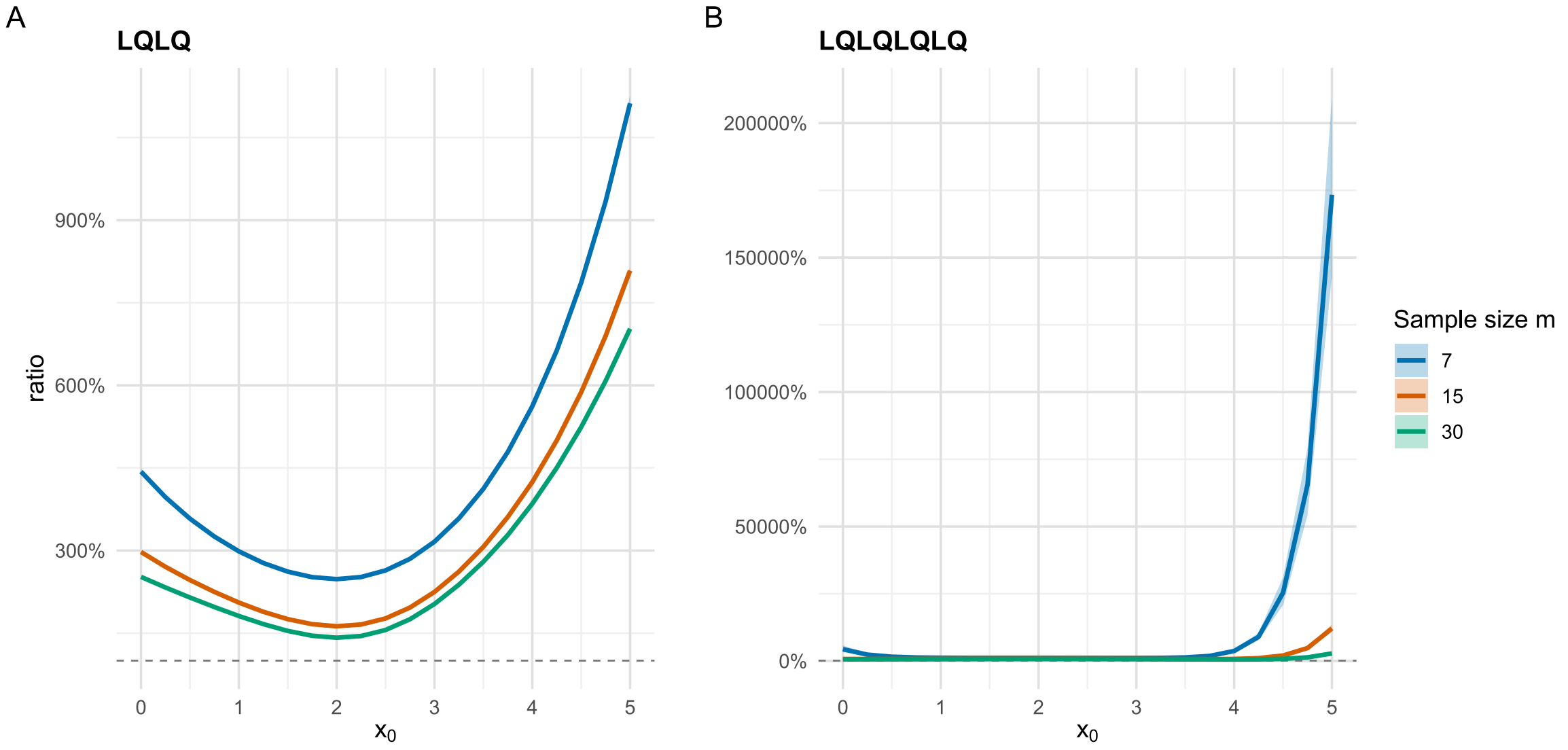

FIGURE S3: MEAN MCSIM INTERVAL-WIDTH RATIOS WITH LATENT-TO-OBSERVED TRAINING. PANELS A AND B SHOW LQLQ AND LQLQLQLQ; COLORS IDENTIFY $m = 7, 15$ AND $30$. WIDTHS ARE NORMALIZED BY THE TRUE 90% NORMAL-INTERVAL WIDTH. RIBBONS ARE POINTWISE 90% OUTER-REPETITION BOOTSTRAP CONFIDENCE INTERVALS. OTHER SIMULATION SETTINGS ARE THE SAME AS IN FIGURE 3.

# S2 Bivariate 3-UO Example and Simulation Results

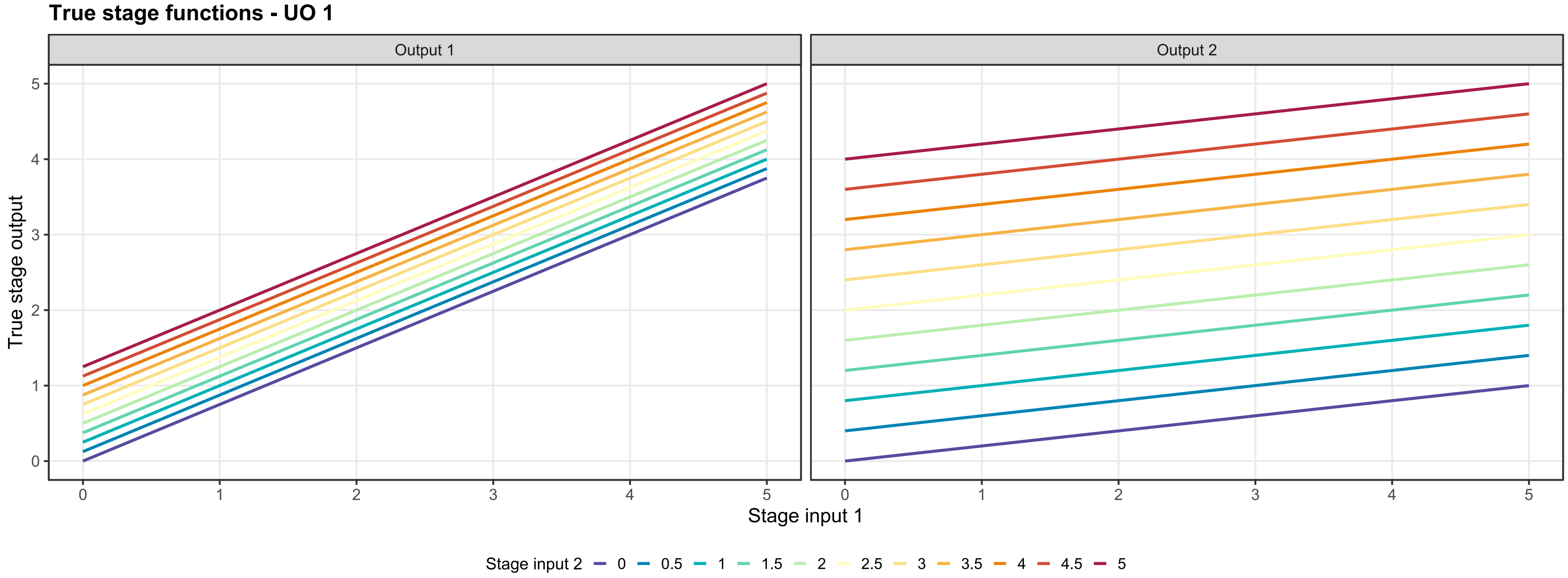

**Figure S4: Function of Unit operation 1 mapping the input variables to the output variables.**

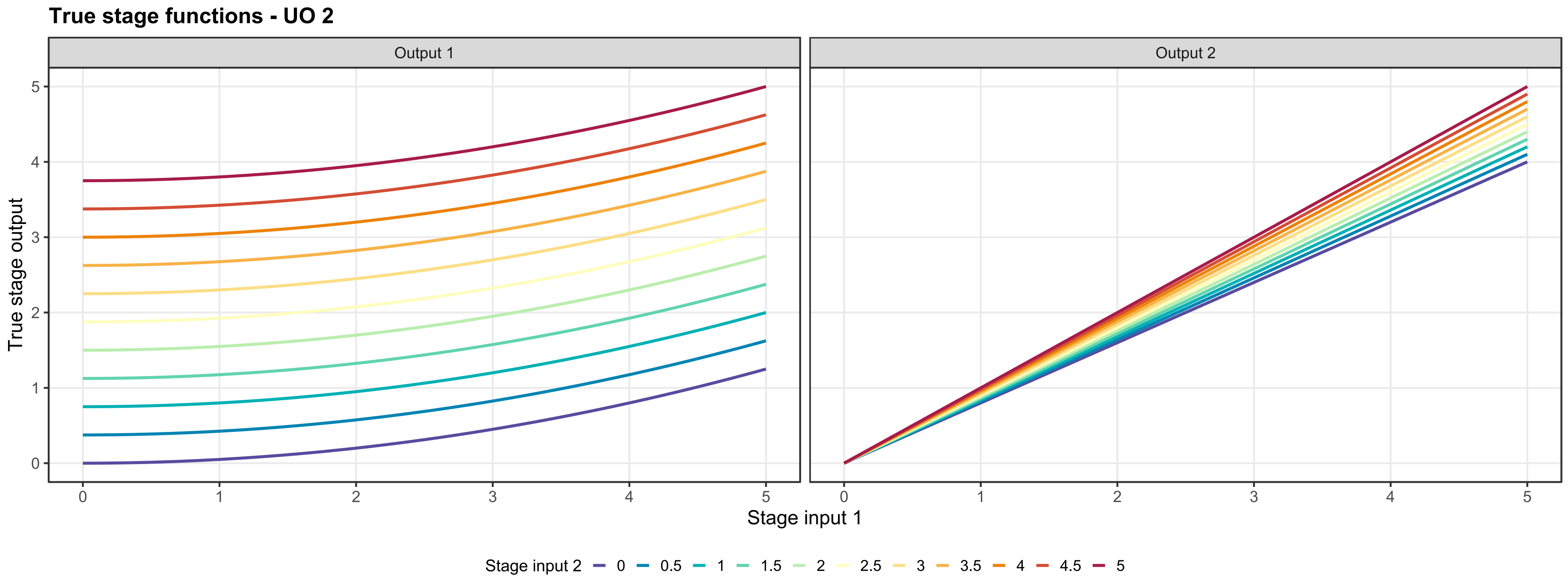

**Figure S5: Function of Unit operation 2 mapping the input variables to the output variables.**

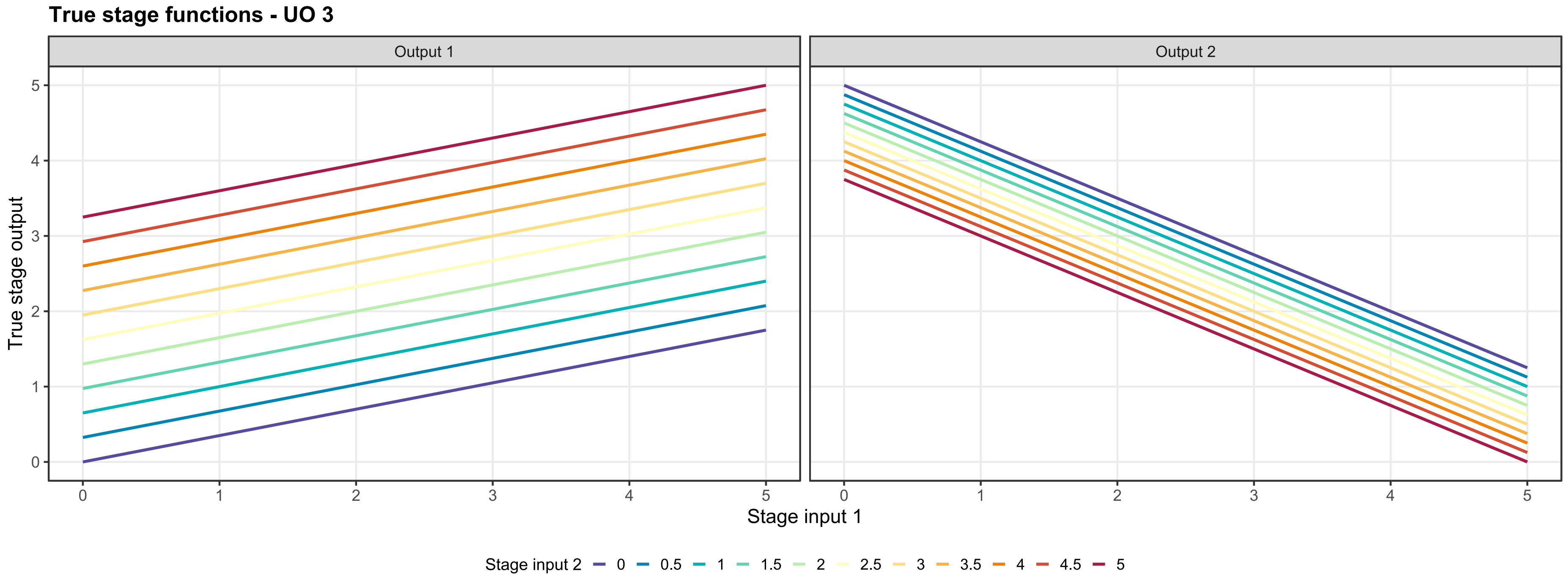

**Figure S6: Function of Unit operation 3 mapping the input variables to the output variables.**

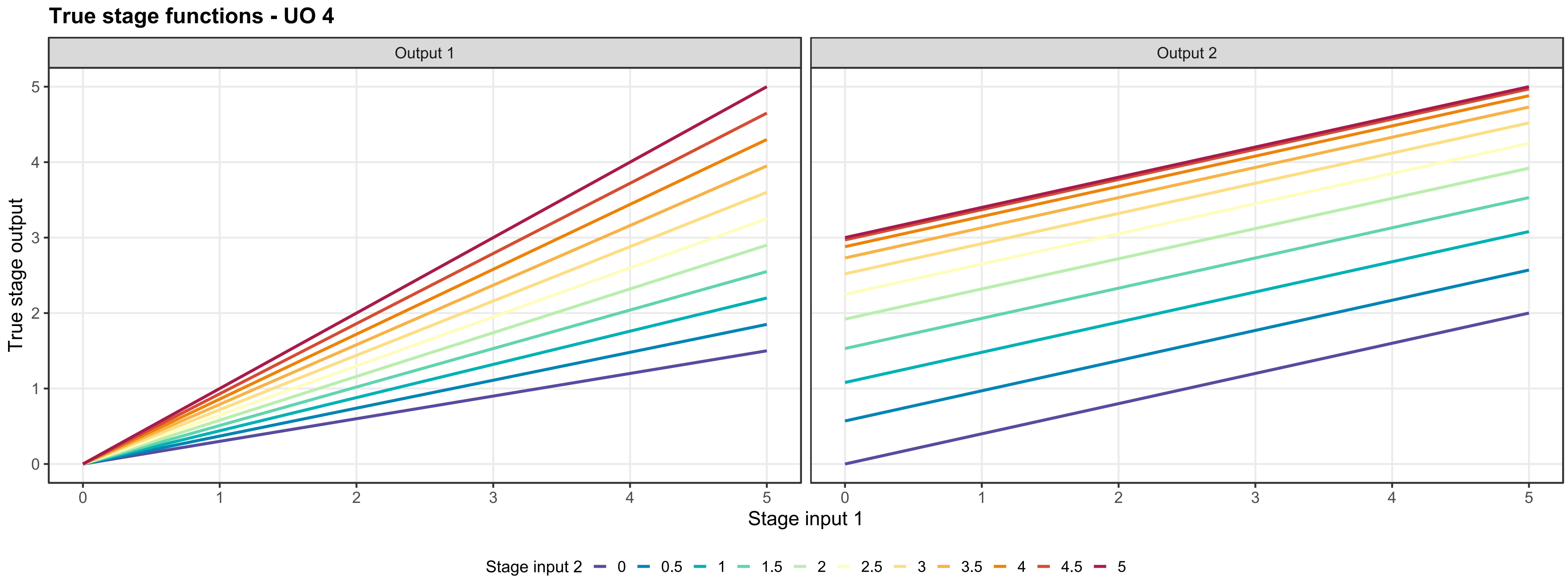

**FIGURE S7: FUNCTION OF UNIT OPERATION 4 MAPPING THE INPUT VARIABLES TO THE OUTPUT VARIABLES.**

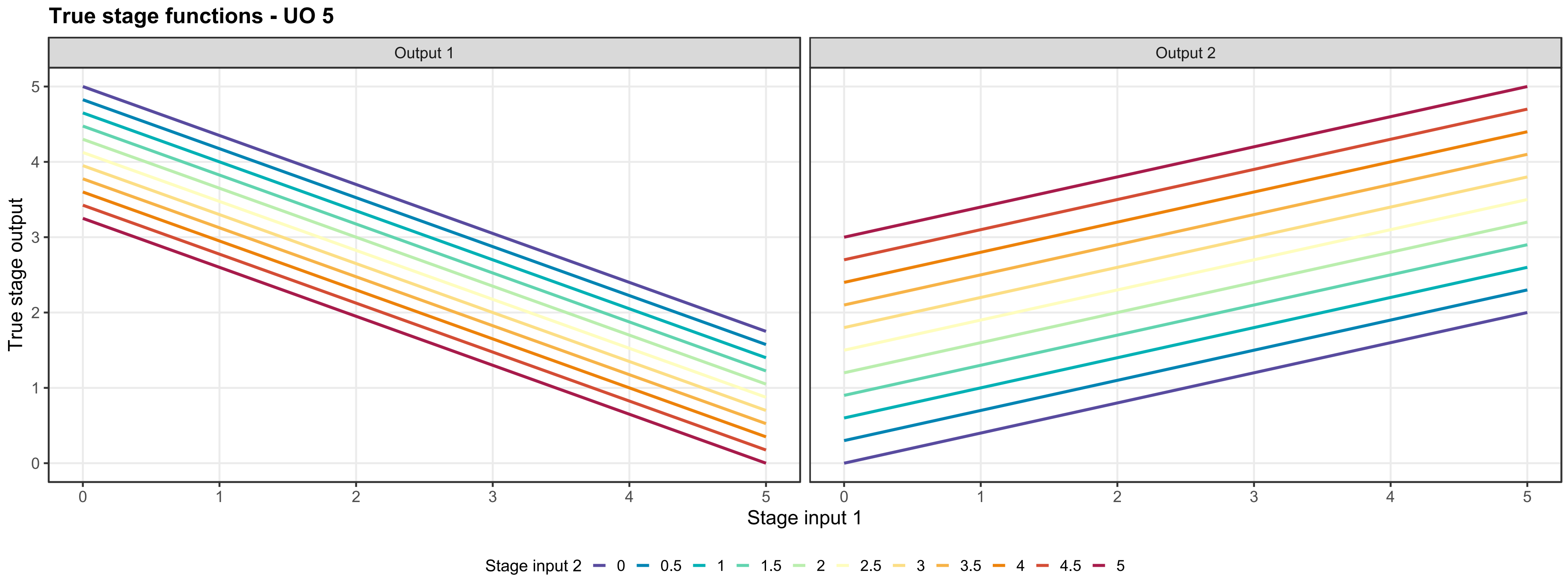

**FIGURE S8: FUNCTION OF UNIT OPERATION 4 MAPPING THE INPUT VARIABLES TO THE OUTPUT VARIABLES.**

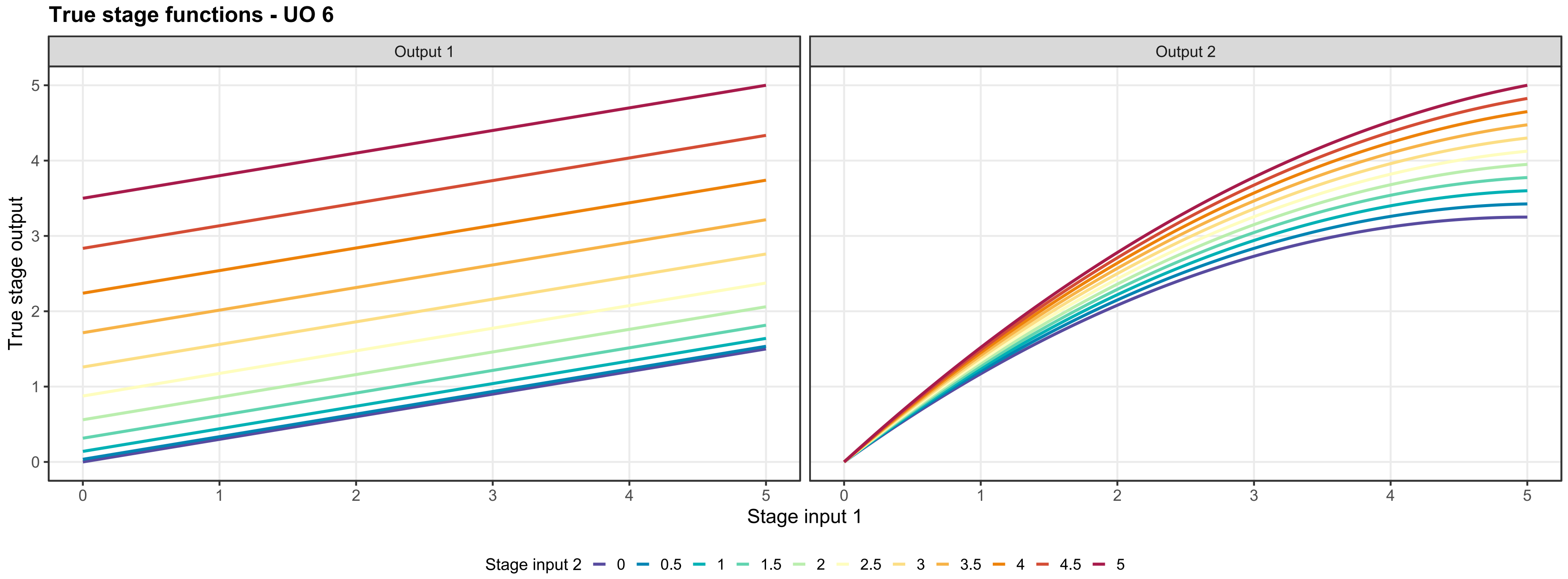

**Figure S9: Function of Unit operation 4 mapping the input variables to the output variables.**

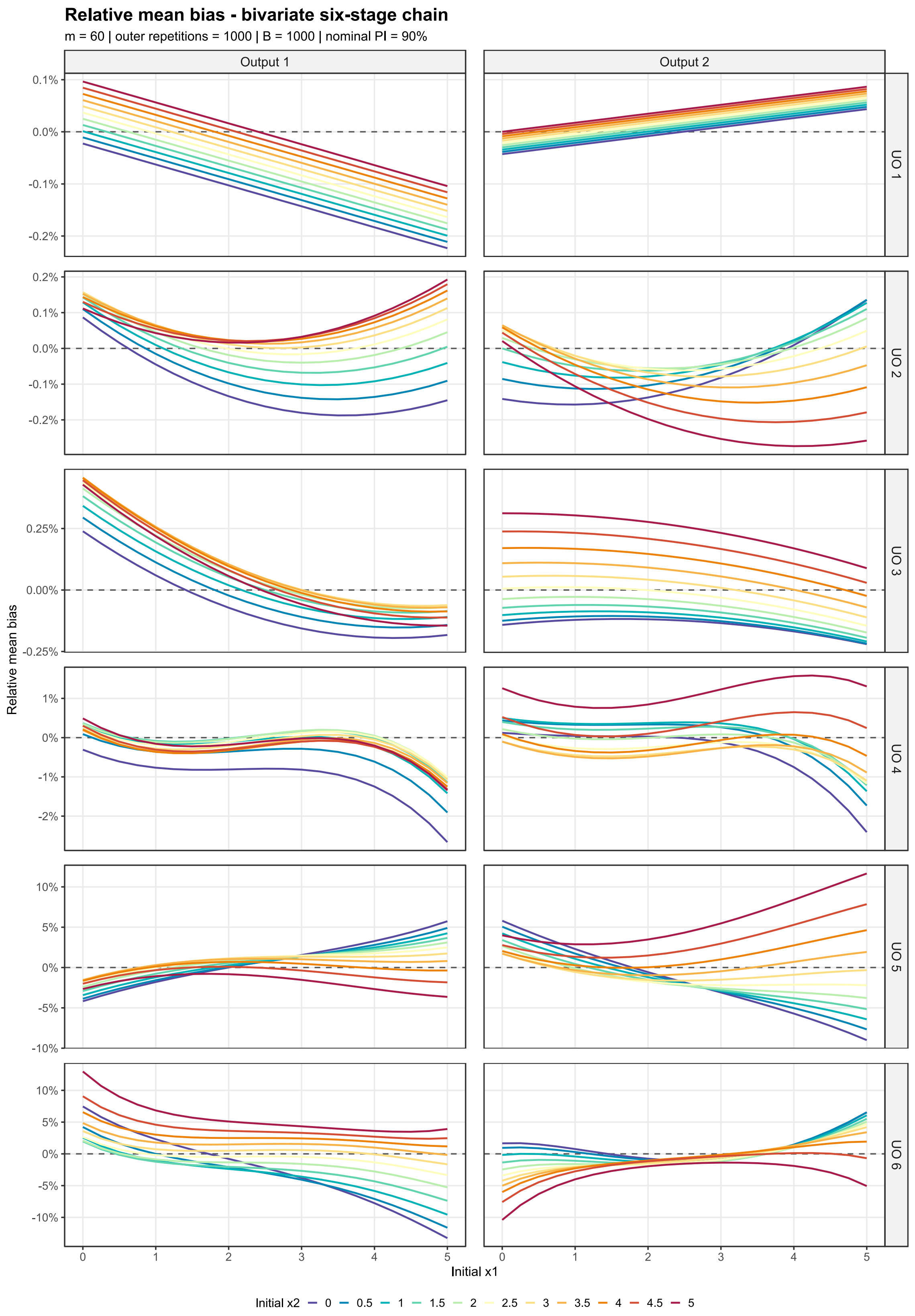

**FIGURE S10: RELATIVE BIAS FOR ALL UOS AND VARIABLES FOR THE APPLICATION OF THE BOOTSTRAP ALGORITHM AND $n = 60$ INPUT DATA POINTS ON AN EQUALLY SPACED GRID OVER THE DESIGN SPACE.**

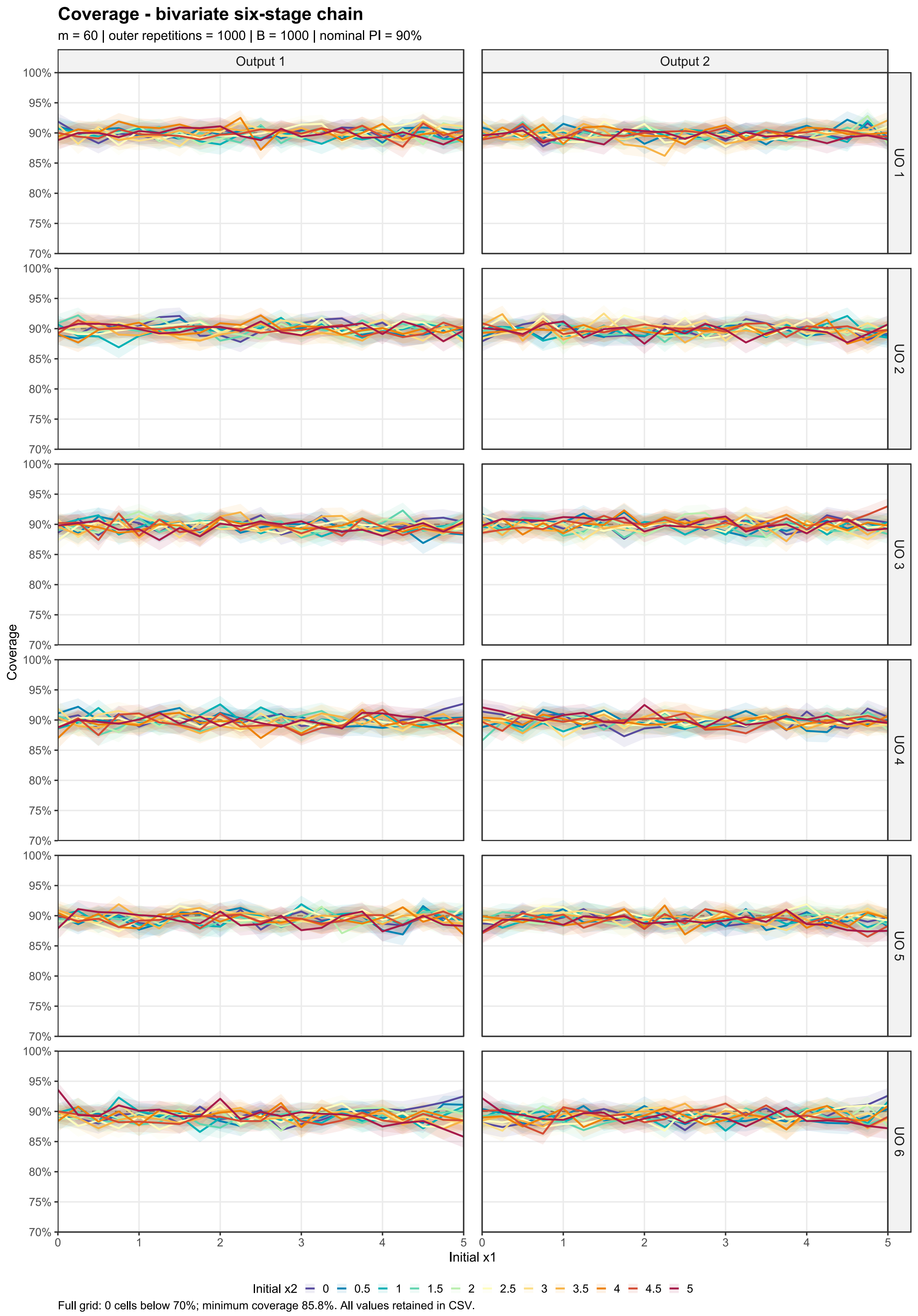

**FIGURE S11: COVERAGE FOR ALL UOS AND VARIABLES FOR THE APPLICATION OF THE BOOTSTRAP ALGORITHM AND $n = 60$ INPUT DATA POINTS.**

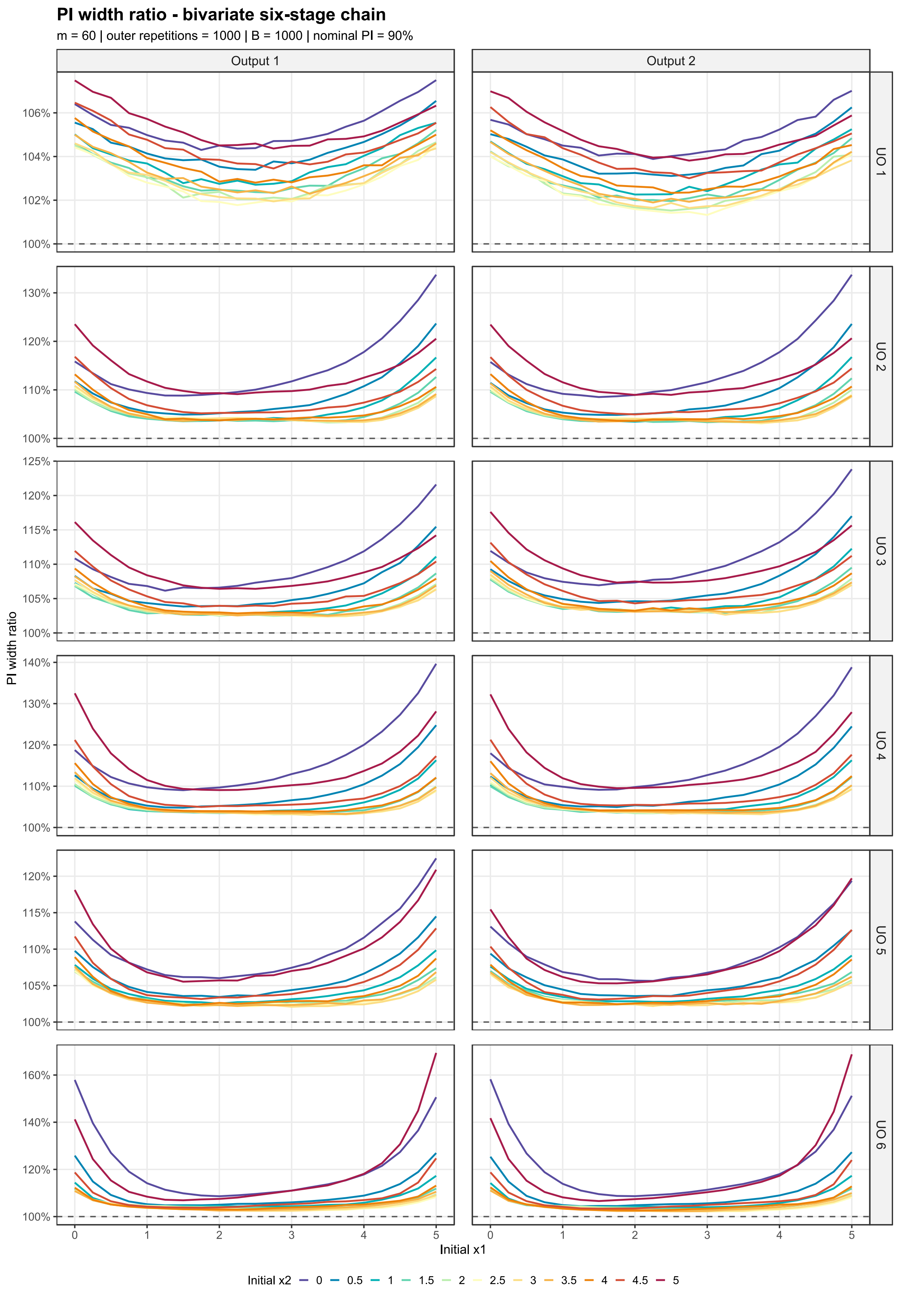

**FIGURE S12: PI WIDTH RATIO FOR ALL UOS AND VARIABLES FOR THE APPLICATION OF THE BOOTSTRAP ALGORITHM AND $n = 60$ INPUT DATA POINTS.**